\documentclass[default,iicol]{sn-jnl}
\usepackage{lmodern}
\usepackage{subfigure}
\usepackage{graphicx}%
\usepackage{multirow}%
\usepackage{amsmath,amssymb}%
\usepackage{amsthm}%
\usepackage[title]{appendix}%
\usepackage[table]{xcolor}%
\usepackage{textcomp}%
\usepackage{manyfoot}%
\usepackage{booktabs}%
\usepackage{algorithm}%
\usepackage{listings}%
\usepackage{graphicx}

\usepackage[switch]{lineno}

\definecolor{Gray}{gray}{0.85}

\usepackage{algorithmic}
\usepackage{array}

\usepackage{booktabs}
\usepackage{graphicx}
\usepackage{bm}
\usepackage{color}
\usepackage{url}
\usepackage{float}
\definecolor{Gray}{gray}{0.85}

\begin{document}

\title[Article Title]{P2Object: Single Point Supervised Object Detection and Instance Segmentation}


\author[1]{\fnm{Pengfei} \sur{Chen}}\email{\{\small{chenpengfei20, yuxuehui17, hanxumeng19, wangkuiran19\}@mails.ucas.ac.cn}}
\equalcont{\small{These authors contributed equally to this work.}}
\author[1]{\fnm{Xuehui} \sur{Yu}}
\equalcont{\small{These authors contributed equally to this work.}}
\author[1]{\fnm{Xumeng} \sur{Han}}

\author[1]{\fnm{Kuiran} \sur{Wang}}

\author[2]{\fnm{Guorong} \sur{Li}}\email{\small{liguorong@ucas.ac.cn}}
\author[3]{\fnm{Lingxi} \sur{Xie}}\email{\small{198808xc@gmail.com}}

\author*[1]{\fnm{Zhenjun} \sur{Han}}\email{\small{hanzhj@ucas.ac.cn}}

\author[1]{\fnm{Jianbin} \sur{Jiao}}\email{\small{jiaojb@ucas.ac.cn}}

\affil[1]{\small{\orgdiv{School of Electronic, Electrical and Communication Engineering}, \orgname{University
of Chinese Academy of Sciences (UCAS)}, \orgaddress{\city{Beijing}, \postcode{101480}, \country{China}}}}

\affil[2]{\small{\orgdiv{School of Computer Science and Technology}, \orgname{University
of Chinese Academy of Sciences (UCAS)}, \orgaddress{\city{Beijing}, \postcode{101480}, \country{China}}}}

\affil[3]{\small{\orgname{Huawei Inc.}, \orgaddress{\city{Shenzhen}, \postcode{518055}, \country{China}}}}

\newcommand{\etal}{\textit{et al}.}
\newcommand{\ie}{\textit{i}.\textit{e}.}
\newcommand{\eg}{\textit{e}.\textit{g}.}
\newcommand{\otsp}{OTSP}


\abstract{Object recognition using single-point supervision has attracted increasing attention recently. However, the performance gap compared with fully-supervised algorithms remains large.
Previous works generated class-agnostic \textbf{\textit{proposals in an image}} offline and then treated mixed candidates as a single bag, 
putting a huge burden on multiple instance learning (MIL).
In this paper, we introduce Point-to-Box Network (P2BNet), which constructs balanced \textbf{\textit{instance-level proposal bags}} by generating proposals in an anchor-like way and refining the proposals in a coarse-to-fine paradigm.
Through further research, we find that the bag of proposals, either at the image level or the instance level, is established on discrete box sampling. This leads the pseudo box estimation into a sub-optimal solution, resulting in the truncation of object boundaries or the excessive inclusion of background. Hence, we conduct a series exploration of discrete-to-continuous optimization, yielding P2BNet++ and Point-to-Mask Network (P2MNet). P2BNet++ conducts an approximately continuous proposal sampling strategy by better utilizing spatial clues. P2MNet further introduces low-level image information to assist in pixel prediction, and a boundary self-prediction is designed to relieve the limitation of the estimated boxes. Benefiting from the continuous object-aware \textbf{\textit{pixel-level perception}}, P2MNet can generate more precise bounding boxes and generalize to segmentation tasks.
Our method largely surpasses the previous methods in terms of the mean average precision on COCO, VOC, SBD, and Cityscapes, demonstrating great potential to bridge the performance gap compared with fully supervised tasks.}





\maketitle
\vspace{-10pt}

\begin{figure*}[htb!]
     \centering
    \includegraphics[width=1.0\linewidth]{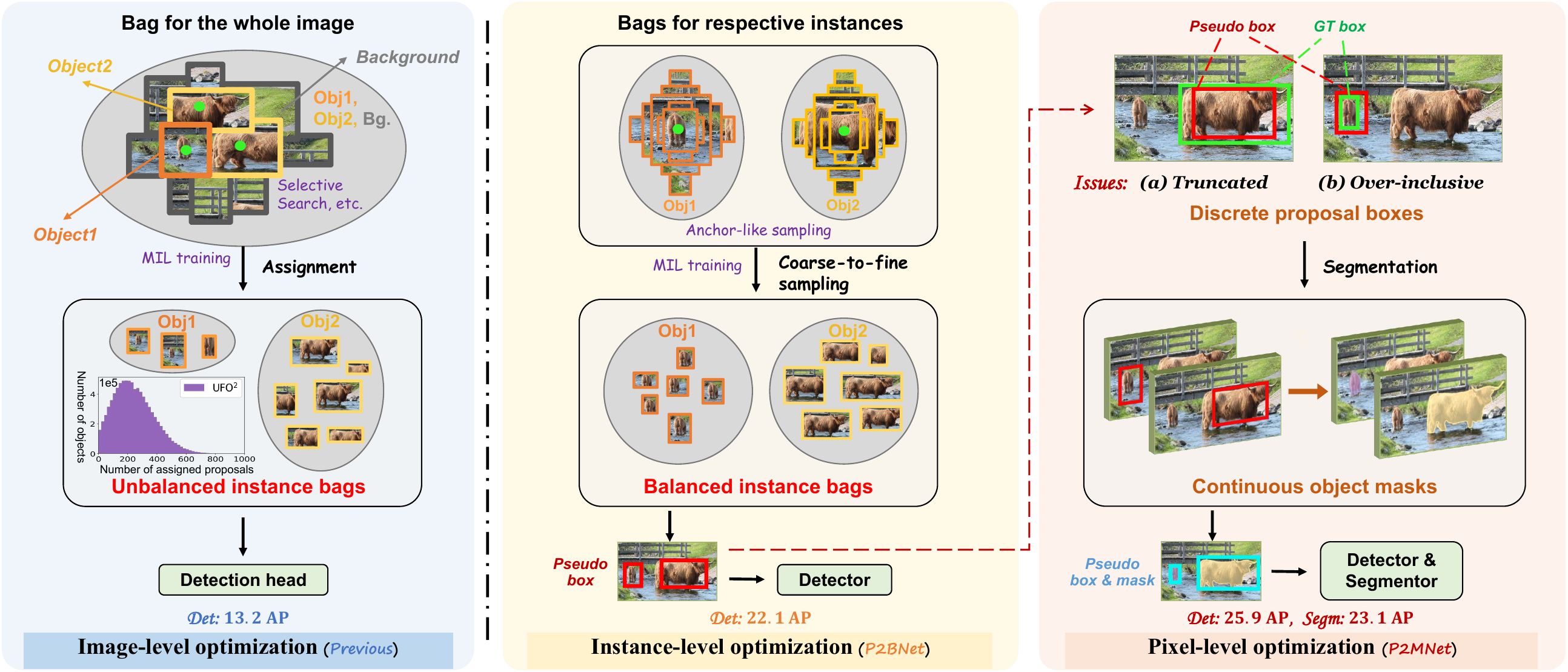}
    
    \caption{
  \textbf{i) Left:} The previous methods, \eg~UFO$^2$, treat all the candidate proposals (like object1, object2 and background) of the image in a single bag, and the numbers of proposals for different are unbalanced. \textbf{ii) Middle:} P2BNet constructs balanced instance-level proposal bags anchor-like sampling and refine the proposals in a coarse-to-fine fashion, which is beneficial for MIL optimization. \textbf{iii) Right:} However, the bag of proposals, either at the image level or instance level, is established on discrete box sampling, which leads to truncated or over-inclusive prediction at the object boundary. P2MNet aims to transfer the optimization to pixel-level and predicts continuous object masks. The estimated pseudo boxes and masks are used to supervise the detector and segmentor, which significantly improves the performance.}
    \vspace{-10pt}
    \label{fig:motivation}
\end{figure*}

\section{Introduction}\label{sec1}

Object detectors ~\cite{FasterRCNN,DBLP:retinanet_focalloss,DBLP:DETR,DBLP:sparsercnn,scalematch,DBLP:Context-Condensation} trained via accurate bounding box supervision have been well received in academia and industry. However, collecting quality bounding box annotations is a labor-intensive task. To solve this problem, weakly supervised object detection~\cite{DBLP:wsddn,DBLP:oicr,DBLP:pcl,DBLP:MELM,DBLP:cam,DBLP:wccn} (WSOD) replaces bounding box annotations with low-cost image-level annotations. Despite the benefits, WSOD methods struggle with complex scenarios and difficulty in distinguishing dense objects. 
Recently, point-based annotation has become increasingly popular in various computer vision tasks, including localization~\cite{CPR,DBLP:withoutboundingbox,p2pnet,DBLP:CPR++}, object detection~\cite{Click,UFO2,DBLP:Group-R-CNN,DBLP:Points_As_Queries,tod_sp}, rotated object detection~\cite{DBLP:pointobb,DBLP:point2rbox,DBLP:p2rbox,DBLP:p-to-rbox,DBLP:pointobb-v2}, instance segmentation~\cite{DBLP:Attnshift,DBLP:BESTIE,DBLP:pointly-sup,DBLP:active_point,DBLB:sapnet}, panoptic segmentation~\cite{DBLP:point_panoptic_seg,DBLP:point2mask} and action localization~\cite{Lee_2021_ICCV}. It provides distinctive location information of the object and is a more cost-effective alternative to bounding box supervision.
However, the performance of point-supervised methods~\cite{Click,UFO2} still lags behind that of fully supervised detectors. 

Previous works ahead offline generate class-agnostic proposals in an image using off-the-shelf proposal (\otsp) methods (\eg, Selective Search~\cite{DBLP:selectivesearch,DBLP:selectivesearch——ob}, MCG~\cite{DBLP:MCG}, and EdgeBox~\cite{DBLP:EdgeBox}), then treat all the mixed candidate proposals into a single bag, 
shown in Fig.~\ref{fig:motivation},
put a huge burden on multiple instance learning (MIL). Besides, the previous work always assigns candidate proposals to different objects by label assignment, which is often inter-object unbalanced (see statistical chart in the first column of Fig.~\ref{fig:motivation}) and can lead to optimization favor towards objects with more candidate proposals, which is not fair to those with fewer proposals.

\begin{figure*}[htb!]
        \centering
        \includegraphics[width=1.\linewidth]{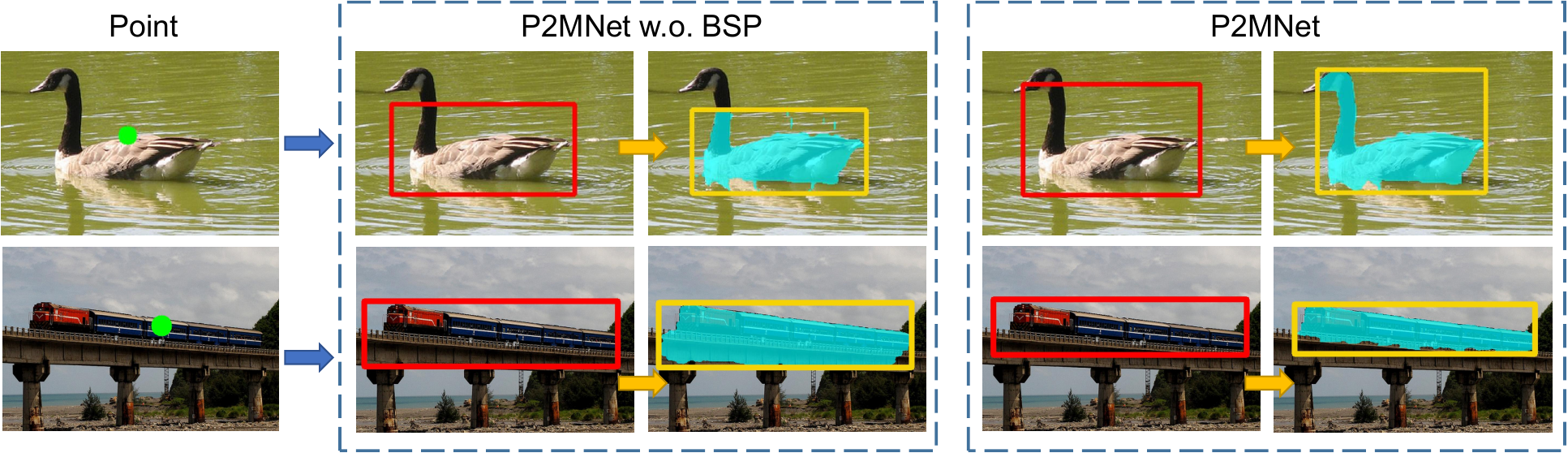}
        \vspace{-10pt}
    \caption{With point annotation, P2BNet++ predicts the pseudo box (red), P2MNet estimates the object mask (blue) and the minimum circumscribed rectangle updates the pseudo box (yellow). Without BSP, the object mask truncates the object influenced by the original pseudo box (first line) or brings much background (second line). With BSP, the above problems are relieved. (Best viewed in color.)}
    \vspace{-15pt}
    \label{fig:Vis_analysis}
\end{figure*}

To address these problems, we present \textbf{P}oint-\textbf{to}-\textbf{B}ox \textbf{Net}work (\textbf{P2BNet}) as a novel approach for constructing intro-object balanced bags for respective instances (the second column of Fig.~\ref{fig:motivation}), which offers an alternative. P2BNet generates an equal number of proposals for each object in an anchor-like way, covering a range of scales and aspect ratios. 
To enhance the quality of the proposals, which finally contributes to the more precise pseudo box, a coarse-to-fine procedure is designed, consisting of two parts: the coarse pseudo-box prediction (CBP) and the precise pseudo-box refinement (PBR). The CBP stage predicts the coarse scale of objects, and the PBR stage iteratively fine-tunes the scale and position. Our P2BNet generates high-quality, balanced proposal bags and ensures the point annotations contribute at all stages (\ie, before, during, and after MIL training). 

Through our further research, we find that the bag of proposals, either at the image level or instance level, is established on discrete box sampling, which cannot fit closely around the target boundary, leading the pseudo box estimation into the sub-optimal solution. Our first attempt is to transfer discrete optimization to the continuous one, yielding \textbf{P2BNet++}. A spatial self-distillation strategy is adopted to process discrete hand-crafted anchors regressed to objects' boundaries when constructing proposal bags, which enhances performance. \textbf{P}oint-\textbf{to}-\textbf{M}ask \textbf{Net}work (\textbf{P2MNet)} is a further exploration of discrete-to-continuous prediction after the P2BNet++. P2BNet++ progresses from discrete anchors to continuous regressed boxes in prediction, while P2MNet further transforms the discrete spatial sampling into continuous pixel perception by combining low-level image information, such as color similarity, with the estimated boxes. Specifically, P2BNet serves as the original pseudo-box predictor, simultaneously predicting a set of dynamic convolution parameters. The object-aware pixel perception (OAPP) module concatenates the feature map and the position embedding of the estimated box, then convolves the feature map with the estimated dynamic convolution parameters for the object maps.
While it still suffers from truncated or over-inclusive prediction at the object boundary. As shown in the third column of Fig.~\ref{fig:Vis_analysis}, the predicted pseudo box \textbf{truncates} a waterbird's neck, or \textbf{over-includes} bridge under the train, encapsulating more background.
The estimated bounding boxes may not cut the objects precisely at the object edge, thus leading to the degradation of box regression in detection training and damaging the performance.
Hence, a boundary self-prediction 
(BSP) module is designed to correct the truncated and overinclusive predictions by weakening the constraints from the inaccurate pseudo boxes at the target boundary, leading to more precise object predictions.
P2MNet can not only generalize to point supervised instance segmentation (PSIS) tasks with estimated object masks but also feedback on the detection results.

Our model's effectiveness and robustness have been demonstrated through detailed experiments on several datasets, including MS COCO~\cite{coco} (2014 \& 2017), Pascal VOC~\cite{DBLP:VOC} (2007 \& 2012), SBD~\cite{DBLP:VOC} and Cityscapes~\cite{cityscape}  Our model has achieved superior performance by a significant margin compared to previous point-based detectors and segmentors. Contributions are summarized as follows:

\begin{itemize}
    \item[—] We propose the point-to-object (P2Object\footnote{The code will be released at \textcolor{magenta}{\url{github.com/ucas-vg/P2BNet}}.}) framework, containing P2BNet and P2MNet, to generate the accurate bounding box and mask of the object from single point annotation.
\item[—] P2BNet++, the first attempt to transfer discrete optimization to the continuous one in object perception, processes discrete
hand-craft anchors (in P2BNet) regressed to objects’ boundaries when constructing proposal bags with
a spatial self-distillation strategy.
    \item[—] P2MNet, the further exploration of discrete-to-continuous perception, not only generalizes to PSIS tasks with estimated object masks
but feedback on the detection results. A boundary self-prediction module, combined with the low-level image information, is proposed to break through the limitation of discrete box sampling.
    \item[—]The detection and segmentation performance of the proposed framework with P2MNet under single quasi-center point supervision improves the mean average precision (AP) of the previous best PSOD and PSIS methods on COCO, VOC, SBD, and Cityscapes datasets, bridging the gap between fully and weakly supervised tasks and achieving comparable performance, especially on AP$_{50}$.
\end{itemize}

{\color{cyan}
The article is extended from our conference paper~\cite{DBLP:P2BNET}. The major difference is summarized as follows. \textcolor{teal}{\textbf{Extension of approaches:}} \textbf{1)} We extend our P2BNet to P2BNet++ with a spatial self-distillation strategy for continuous pseudo box generation (Sec.~\ref{sec:p2b++}). \textbf{2)} We give the description of P2MNet (Sec.~\ref{sec:p2mnet}) and demonstrate the structure of P2MNet, which transfers degree box-level sampling to continuous pixel-level perception (Sec.~\ref{sec:oapp}) to enhance the accuracy of the predicted pseudo box. \textbf{3)} We find that the performance of the segmentation branch heavily relies on the quality of pseudo-labels generated by P2BNet or P2BNet++, so the boundary self-prediction module is proposed to relieve truncated and overinclusive prediction problems (Sec.~\ref{sec:BSP}). \textcolor{teal}{\textbf{Extension of the segmentation task:}} \textbf{4)} P2MNet is generalized to point supervised instance segmentation task (Sec.~\ref{sec:PSIS}) with detailed experiments and analyses. \textcolor{teal}{\textbf{Extension of datasets and metrics:}}
\textbf{5)} Experiments on more datasets (VOC, SBD, and Cityscapes) in Sec.~\ref{sec:psod} and Sec.~\ref{sec:PSIS} demonstrate the generalizability of our methods in detection and segmentation tasks. Detailed performance of different metrics is reported.}

\section{Related Work}
\label{RelatedWork}
This section briefly discusses the status of mask-level, box-level, image-level, and point-level supervised object detection and instance segmentation.


\begin{figure*}
    \centering
    \includegraphics[width=1.0\linewidth]{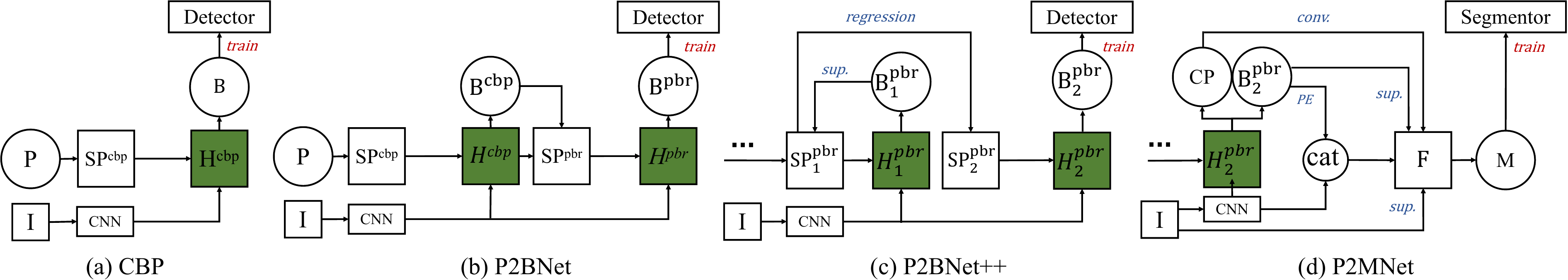}
    \caption{The general pipelines for the proposed CBP, P2BNet, P2BNet++ and P2MNet, respectively. `I' is the input image, `P' is the single point annotation, `B' is the pseudo box, and `M' is the pseudo mask. `SP' is the sampling, `H' is the MIL network, `PE' is position embedding, and `CP' is the convolution parameter. The `cat' is the `concat' operation, `sup.' is the `supervise' operation, and `conv.' is the `convolution' operation.}
    \label{fig:pipeline}
\end{figure*}

\subsection{Object Detection}
\textbf{Box-supervised object detection} \cite{fastrcnn,FasterRCNN,DBLP:yolo,DBLP:SSD,DBLP:retinanet_focalloss,DBLP:DETR,DBLP:sparsercnn,scalematch, DBLP:Context-Condensation} gives the network a specific category and box information. 
 Proposal-based detectors use \otsp~methods (like the selective search~\cite{DBLP:selectivesearch} in Fast R-CNN~\cite{fastrcnn}) or deep networks (like RPN in Faster R-CNN~\cite{FasterRCNN}) to predict proposals and then sparsely make classifications. 
 Anchor-based detectors~\cite{DBLP:yolo,DBLP:SSD, DBLP:retinanet_focalloss} directly predict proposals using anchors. 
Recently, query-based detectors (DETR~\cite{DBLP:DETR}, Deformable-DETR~\cite{DBLP:DeformableDETR}, Sparse R-CNN~\cite{DBLP:sparsercnn}, and DINO~\cite{DBLP:DINO}) utilize global information for better representation. 
Grounding DINO~\cite{DBLP:GROUNDINGDINO}, a CLIP~\cite{DBLP:CLIP}-based detector trained on large-scale datasets, can detect each category with the text of class name input.
However, they require high human costs.


\noindent\textbf{Image-supervised object detection}~\cite{DBLP:wsddn,DBLP:oicr,DBLP:pcl,DBLP:slv,DBLP:MELM,DBLP:cam,DBLP:wccn,DBLP:Acol,DBLP:weaklysurvey,DBLP:uwsod} is a traditional field in WSOD.
In CAM-based methods~\cite{DBLP:cam,DBLP:wccn,DBLP:Acol}, the main purpose is to produce the class activation maps (CAM)~\cite{DBLP:cam}, use a threshold to choose a high-score region and find the largest general domain. 
In MIL-based methods~\cite{DBLP:wsddn,DBLP:oicr,DBLP:pcl,DBLP:slv,DBLP:MELM}, a bag is positively labeled if it contains at least one positive instance; otherwise, it is negative. 
WSDDN~\cite{DBLP:wsddn} introduced MIL into WSOD with a representative two-stream weakly supervised deep detection network that can classify positive proposals. OICR and PCL~\cite{DBLP:oicr, DBLP:pcl} introduce an iterative fashion into WSOD and attempt to find the whole instead of a discriminative part. W2N~\cite{DBLP:W2N} switches weak supervision to noisy supervision, SPE~\cite{SPE2022} uses sparse proposal evolution.

\noindent\textbf{Point-supervised object detection} annotation is a relatively recent innovation. It bridges the gap between fully-supervised tasks and image-supervised vision tasks.  
\cite{DBLP:Group-R-CNN,DBLP:Points_As_Queries} use a certain percentage of box-level labels and point annotations to supervise the detectors in a semi-supervised fashion. For single point supervision setting, \cite{Click} introduces center-click annotation to replace box supervision and estimates a scale with the error between two times of center-click. UFO$^{2}$~\cite{UFO2} design a network compatible with various supervision forms, such as tags, points, scribbles, and box annotation. However, these frameworks are based on \otsp~methods and are not specially designed for point annotation. 
Therefore, their performance is limited, and they perform poorly in complex scenarios, like the COCO~\cite{coco} dataset. 
We introduce a new framework with P2BNet, which is free of \otsp~methods. Pixel perception extends P2BNet into the continuous object representation.

\subsection{Instance Segmentation}
\label{sec:survey_of_seg}
\textbf{Mask-supervised instance segmentation}~\cite{DBLP:MASK-RCNN,DBLP:BlendMask,DBLP:Mask_Encoding,DBLP:SOLO,DBLP:SOLOv2,DBLP:condinst,DBLP:mask2former,DBLP:conf/nips/WangLL22a,DBLP:maskdino} requires mask and category annotation for supervision. The most widely used approach is a two-stage method, \eg, Mask R-CNN~\cite{DBLP:MASK-RCNN}, which predicts bounding boxes first and then estimates the object mask for each bounding box. Recently, ~\cite{DBLP:BlendMask,DBLP:Mask_Encoding} built their methods on anchor-free object detectors~\cite{DBLP:FCOS}. CondInst~\cite{DBLP:condinst} frees object masks from the RoI level to the whole feature map level, which is of high resolution. SOLO v1\&2~\cite{DBLP:SOLO, DBLP:SOLOv2, DBLP:condinst} use two-dimensional coordinate embedding and directly predict object masks without bounding boxes. Mask2former~\cite{DBLP:mask2former} adopts transformer structure into instance segmentation.~\cite{DBLP:conf/nips/WangLL22a} devises a new training framework that boosts query-based models through discriminative query embedding learning. Mask DINO~\cite{DBLP:maskdino} adopts a denoising strategy to make the optimization more stable.


\noindent\textbf{Weakly-supervised instance segmentation} (WSIS)~\cite{DBLP:IRNet,PDSL,DBLP:BESTIE,DBLP:Boxinst} aims to discover an object at mask-level with weak supervision. With image-level supervision, IRNet\cite{DBLP:IRNet} obtains pseudo instance segmentation labels by learning class-agnostic instance maps. PDSL\cite{PDSL} integrates bottom-up object cues into the top-down pipeline and proposes a learning framework for unified parallel detection and segmentation. BESTIE~\cite{DBLP:BESTIE} transfers semantic knowledge into instance segmentation and then makes refinement. ~\cite{DBLP:BBTP,DBLP:Boxinst} conduct segmentation under box-level supervision (BSIS). BoxInst~\cite{DBLP:Boxinst} conducts detection and then utilizes mask projection and low-level color information to achieve box-supervised segmentation.

\begin{figure*}[tb!]
    \begin{center}
    \includegraphics[width=1.\linewidth]{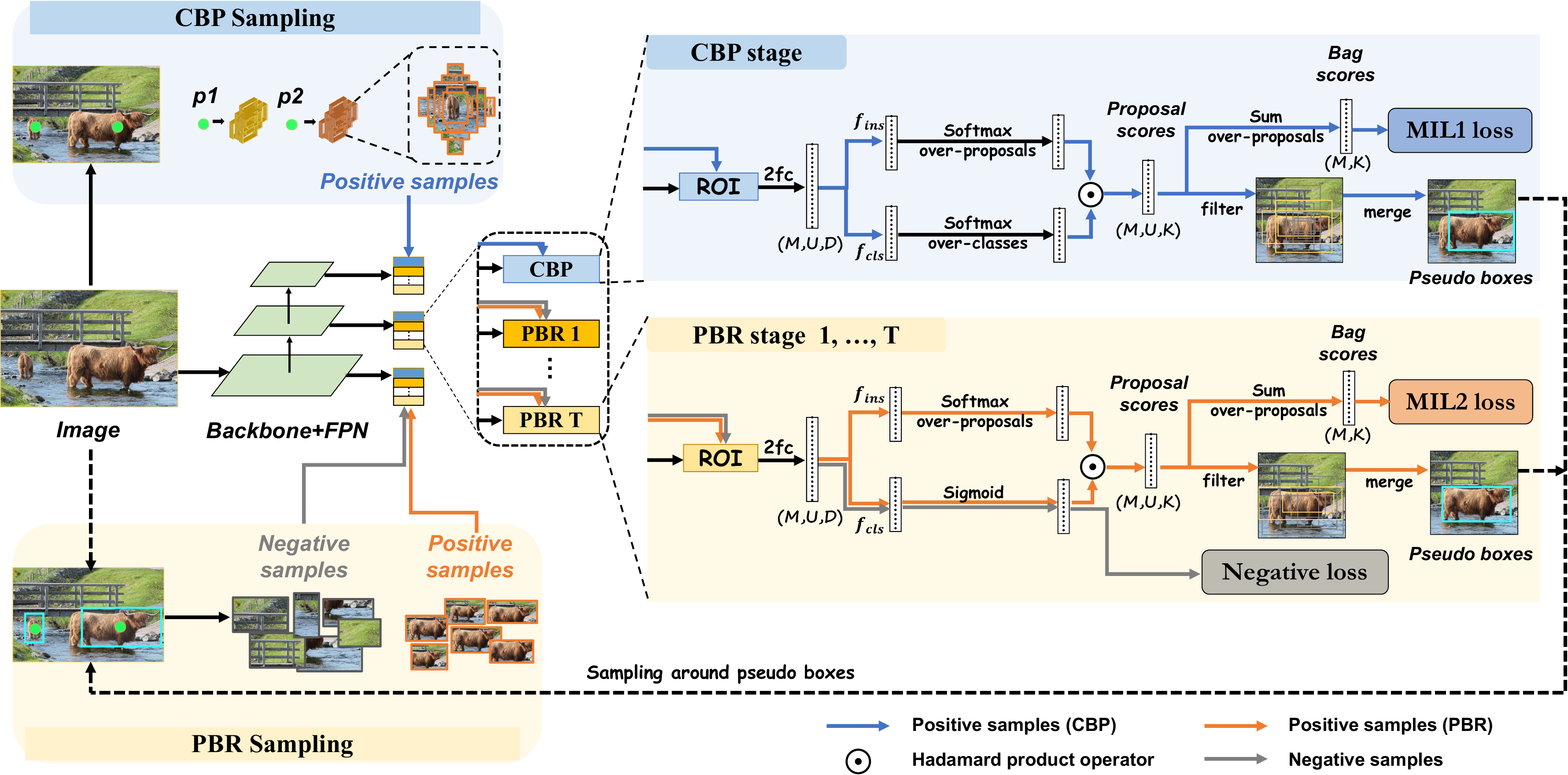}
    \vspace{-5pt}
    \caption{The architecture of P2BNet. Firstly, proposal bags are fixedly sampled around point annotations for classifier training to predict coarse pseudo boxes in the CBP stage. Secondly, to predict refined pseudo boxes in the PBR stage, high-quality proposal bags and negative proposals are sampled with coarse pseudo boxes for training. Finally, the pseudo boxes train a classic detector. (Best viewed in colour.)}
    \vspace{-13pt}
    \label{fig:framework}
    \end{center}
\end{figure*}
\noindent\textbf{Point-supervised instance segmentation}~\cite{DBLP:pointly-sup, DBLP:active_point,DBLP:point2mask, DBLP:journals/ijcv/FanZ23,DBLP:point_panoptic_seg,DBLP:WISE-Net,DBLP:BESTIE,DBLP:Attnshift}
follows a trade-off style, which is time-efficient but high-performance. 
Recent methods~\cite{DBLP:pointly-sup, DBLP:active_point} uses sever points and bounding boxes to supervise instance segmentation. ~\cite{DBLP:point2mask, DBLP:point_panoptic_seg} uses single point to supervise panoptic segmentation.
Some single-point supervised instance segmentation methods, such as WISE-Net~\cite{DBLP:WISE-Net}, handle point-level labels with well-done proposals. WISE-Net first uses L-Net to localize the object. Then, it groups pixels in the image to select the mask proposal with the highest degree of matching with E-Net. BESTIE~\cite{DBLP:BESTIE} utilizes point supervision to produce more refined mask proposals by passing instance-level cues to WSSS, but it fails to utilize the full potential of point-based annotations. Attnshift~\cite{DBLP:Attnshift} shifts the annotated point to several points, estimates the part regions and combines the parts into the whole object as the mask output. 

\section{Methodology}
The P2Object framework, which includes P2BNet (or P2BNet++) and P2MNet, yields accurate object bounding boxes and masks from single-point annotations. Its progressive pipeline is shown in Fig.~\ref{fig:pipeline}. Lacking ground-truth box and mask/contour annotations, object detection and segmentation shift from a fitting problem to an estimation problem. In this paper, P2Object represents a significant step forward, transitioning from discrete to continuous estimation. \textbf{CBP stage}, the coarse prediction stage of P2BNet, samples proposals with hand-crafted anchors and selects the pseudo box via MIL classifier for detector training. \textbf{PBR stage}, for refinement, builds a better proposal set with denser anchors based on CBP output. With negative samples, the PBR enhances CBP learning by preventing it from falling into sub-optimality and predicts a more precise pseudo box. \textbf{P2BNet++} utilizes the estimated pseudo box to guide hand-crafted anchor regression, thereby enhancing the quality of proposal bags and pioneering the transition from discrete to continuous estimation in object perception. Building upon the advancements of P2BNet++, \textbf{P2MNet} further explores this conversion. It estimates the pseudo box (as a position embedding) and the convolution parameter (CP), with the CP operating on the feature map to predict the object mask. The pseudo-box and pixel similarity supervise the mask prediction process. A boundary self-prediction module corrects inaccurate predictions by relaxing constraints on low-quality pseudo-boxes at boundaries, thereby enhancing precision. P2MNet not only extends to PSIS tasks with estimated masks but also provides feedback for detection results.

\subsection{Point-to-Box Network}
The P2BNet-FR framework comprises two components: the Point-to-Box Network (P2BNet) and Faster R-CNN (FR). P2BNet predicts pseudo boxes with point annotations to train the detector, and we use standard settings for FR without any bells and whistles.

The architecture of P2BNet is shown in Fig.~\ref{fig:framework}. It includes the \textbf{C}oarse Pseudo \textbf{B}ox \textbf{P}rediction (CBP) stage and the \textbf{P}seudo \textbf{B}ox \textbf{R}efinement (PBR) stage. 
The CBP stage predicts the object's coarse scale (width and height), and the PBR stage iteratively finetunes the scale and position.
The overall loss function is:

\begin{equation}
\begin{aligned}
\mathcal{L}_{p2b}= \mathcal{L}_{cbp} + \sum\limits_{t=1}^{T} \mathcal{L}_{pbr}^{(t)},
\label{Eq:P2B_loss_all}
\end{aligned}
\end{equation}
where $T$  is the iteration number of PBR stages.

\subsubsection{Coarse Pseudo Box Prediction}
In the CBP stage, taking the annotated point as the box center, proposal boxes of different widths and heights are first generated in an anchor style for each object. Features of the sampled proposals are extracted to train a MIL classifier for selecting the best-fitted proposals.
the top-$k$ merging policy is used to estimate coarse pseudo boxes.

\noindent\textbf{CBP Sampling.} CBP Sampling fixes samples around the annotated point.
With the point annotation $\mathbf{p}=(\mathbf{p}_x,\mathbf{p}_y)$ as the center, $s$ as the size, and $v$ to adjust the aspect ratio, the proposal box $\mathbf{b}=(\mathbf{b}_x, \mathbf{b}_y, \mathbf{b}_w, \mathbf{b}_h)$ is generated, \ie,~$\mathbf{b}=(\mathbf{p}_x, \mathbf{p}_y, v\cdot s, \frac{1}{v} \cdot s)$. The proposal box sampling schematic diagram is shown in Fig.~\ref{fig:sample_cbp_pbr} (Left). 
Through the adjustment of $s$ and $v$, each point annotation $\mathbf{p}_j$ generates a bag of proposal boxes with different manually set scales and aspect ratios (like anchors), denoted by $\mathcal{B}_j$ ($j \in \{1,2,\dots, M \}$, where $M$ is the number of objects). 

\noindent\textbf{CBP Module.} For a proposal bag $\mathcal{B}_j$, features $\mathbf{F}_j\in \mathbb{R}^{U \times D}$ of proposals are extracted through $7 \times 7$ RoIAlign~\cite{DBLP:MASK-RCNN} and two fully connected (fc) layers, where $U$ is the amount of proposals in $\mathcal{B}_j$, and $D$ is the feature dimension. Motivated by WSDDN~\cite{DBLP:wsddn}, we design a two-stream structure as a MIL classifier to find the best bounding box region for object representation. Specifically, applying the classification branch $f_{cls}$ to $\mathbf{F}_j$ yields $\mathbf{O}_{j}^{cls}\in \mathbb{R}^{U \times K}$. It is then passed through the activation function to obtain the classification score $\mathbf{S}^{cls}_j\in \mathbb{R}^{U \times K}$, where $K$ denotes the number of instance categories. Likewise, instance score $\mathbf{S}^{ins}_j\in \mathbb{R}^{U \times K}$ is obtained through instance selection branch $f_{ins}$ and activation function, \ie, 
\begin{equation}
\setlength\abovedisplayskip{2pt}
\setlength\belowdisplayskip{0pt}
[\mathbf{S}^{cls}_j]_{uk} = e^{[\mathbf{O}^{cls}_j]_{uk}}\big/\sum\limits_{i=1}^{K} e^{[\mathbf{O}^{cls}_j]_{ui}}, \label{Eq:S_cls_branch}
\end{equation}
\begin{equation}
\setlength\abovedisplayskip{0pt}
 [\mathbf{S}^{ins}_j]_{uk} = e^{[\mathbf{O}^{ins}_j]_{uk}}\big/\sum\limits_{i=1}^U e^{[\mathbf{O}^{ins}_j]_{ik}},
\label{Eq:S_ins_branch}
\end{equation}
\noindent where $[\cdot]_{uk}$ denotes the value at row $u$ and column $k$ in the matrix. We obtain the proposal score $\mathbf{S}_j$ by computing the Hadamard product of the classification score and the instance score, and the bag score $\widehat{\mathbf{S}}_j$ is obtained through the summation of the proposal scores of $U$ proposal boxes, \ie,
\begin{equation}
\begin{aligned}
\setlength\abovedisplayskip{1pt}
\setlength\belowdisplayskip{1pt}
\mathbf{S}_j=\mathbf{S}^{cls}_j \odot \mathbf{S}^{ins}_j \in \mathbb{R}^{U \times K}, \quad
\widehat{\mathbf{S}}_j & = \sum\limits_{u=1}^{U} [\mathbf{S}_j]_u \in \mathbb{R}^{K}.
\label{Eq:S_bag2}
\end{aligned}
\end{equation}
\noindent $\widehat{\mathbf{S}}_j$ can be viewed as the weighted summation of the classification score $[\mathbf{S}^{cls}_j]_u$ by the corresponding selection score $[\mathbf{S}^{ins}_j]_u$.

\noindent\textbf{CBP Loss.}
The MIL loss in the CBP module (termed $\mathcal{L}_{mil1}$ to distinguish it from the MIL loss in PBR) uses the form of the cross-entropy loss, defined as: 
\begin{equation}
\setlength\abovedisplayskip{3pt}
\setlength\belowdisplayskip{3pt}
\begin{aligned}
\mathcal{L}_{cbp} = \mathrm{\alpha_{1}}&\mathcal{L}_{mil1} =-\frac{\mathrm{\alpha_{1}}}{M}\sum\limits_{j=1}^{M} \sum\limits_{k=1}^{K} [\mathbf{c}_j]_k 
\log([\widehat{\mathbf{S}}_j]_k)\\
&+ (1-[\mathbf{c}_j]_k)  \log(1-[\widehat{\mathbf{S}}_j]_k),
\label{Eq:bce loss}
\end{aligned} 
\end{equation}where $\mathbf{c}_{j} \in \{0, 1\}^{K}$ is the one-hot category label, and $\alpha_{1}$ is 0.25.
The CBP loss is to make each proposal correctly predict the category and the instance it belongs to. 

Finally, the top-$k$ boxes with the highest proposal score $\mathbf{S}_j$ are weighted to obtain coarse pseudo boxes for the following PBR sampling. 

\subsubsection{Pseudo Box Refinement}
The PBR stage aims to finetune the position, width and height of pseudo boxes and can be iteratively performed in a cascaded fashion to improve performance. By adjusting the height and width of the pseudo box obtained in the previous stage (or iteration) in a small span while jittering its center position, finer proposal boxes are generated as positive examples for module training. Further, since positive proposal bags are generated in the local region, negative samples can be sampled far from the proposal bags to suppress the background. PBR module weights the top-$k$ proposals with the highest predicted scores as the output.

\begin{figure*}[tb!]
    \centering
    \includegraphics[width=1.\linewidth]{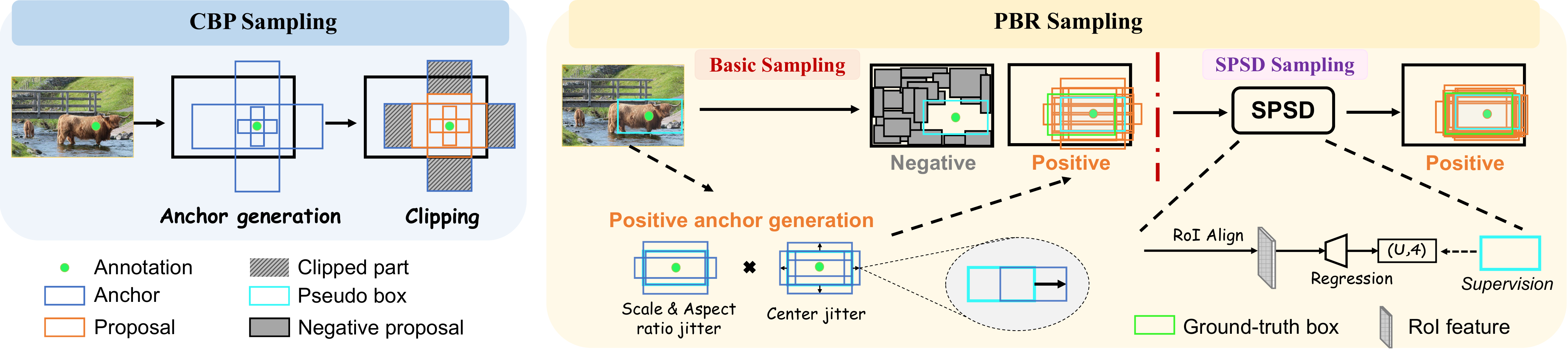}

    \caption{Details of sampling strategies in the CBP and PBR stages. The arrows in PBR sampling mean the offset of center jitter. The basic sampling is conducted through the center jitter following scale and aspect ratio jitter. The SPSD sampling, adopting spatial distillation policy, is used in P2BNet++.}
    \label{fig:sample_cbp_pbr}
    \vspace{-10pt}
\end{figure*}

\noindent\textbf{PBR Sampling.}
PBR basic sampling adapts samples around estimated boxes. As shown in Fig.~\ref{fig:sample_cbp_pbr} (Right), for each coarse pseudo box $b^*=(b^*_x,b^*_y,b^*_w,b^*_h)$ obtained in the previous stage (or iteration), we adjust its scale and aspect ratio with $s$ and $v$ manually and jitter its position with $\mathbf{o}=(\mathbf{o}_x, \mathbf{o}_y)$ (move it in all directions and let the network learn on its own) to obtain the finer proposal $\mathbf{b}=(\mathbf{b}_x, \mathbf{b}_y, \mathbf{b}_w, \mathbf{b}_h)$:
\begin{equation}
\setlength\abovedisplayskip{2pt}
\setlength\belowdisplayskip{0pt}
\mathbf{b}_w = v \cdot s \cdot \mathbf{b}^*_w, \quad \mathbf{b}_h = \frac{1}{v} \cdot s \cdot \mathbf{b}^*_h, 
\setlength\abovedisplayskip{0pt}
\label{Eq: PRB sampling}
\end{equation}
\begin{equation}
\setlength\abovedisplayskip{2pt}
\setlength\belowdisplayskip{0pt}
\setlength\abovedisplayskip{0pt}
\mathbf{b}_x = \mathbf{b}^*_x + \mathbf{b}_w \cdot \mathbf{o}_x, \quad \mathbf{b}_y = \mathbf{b}^*_y + \mathbf{b}_h \cdot \mathbf{o}_y.
\label{Eq: PRB sampling 2}
\end{equation}
These finer proposals are used as positive proposal bags $\mathcal{B}_j$ to train the PBR module.
2) \textbf{Negative sampling} is introduced in the PBR stage for better background suppression. 
To compose the negative sample set $\mathcal{NG}$ for the PBR module, we randomly sample many proposal boxes with small IoU (by default set as smaller than 0.3) with all positive proposals in all bags. 

\noindent\textbf{PBR Module.} 
The PBR module has a similar structure to the CBP module. It shares the backbone network and two fully connected layers with CBP. It also has a classification branch $f_{cls}$ and an instance selection branch $f_{ins}$. Note that $f_{cls}$ and $f_{ins}$ do not share parameters between different stages and iterations.
For an instance selection branch, we adopt the same structure as the CBP module and utilize Eq.~\ref{Eq:S_ins_branch} to predict the instance score $\mathbf{S}^{ins}_j $ for the proposal bag $\mathcal{B}_j$. Differently, the classification branch uses the $sigmoid$ activation function $\sigma(x)$ to predict the classification score $\mathbf{S}^{cls}_j $ , \ie,
\begin{equation}
\begin{aligned}
\sigma(x) = 1/(1+e^{-x}), \quad
\mathbf{S}^{cls}_j = \sigma(f_{cls}(\mathbf{F}_j)) \in \mathbb{R}^{U \times K}.
\label{Eq:cls_pbr}
\end{aligned}    
\end{equation}
This form makes multi-label classification possible, which can distinguish overlapping proposal boxes from different objects. According to Eq.~\ref{Eq:S_bag2}, we can calculate bag score $\mathbf{\widehat{S}}^*_j$ by using $\mathbf{S}^{cls}_j$ and $\mathbf{S}^{ins}_j$ of the current stage. 

For the negative sample set $\mathcal{NG}$, we calculate its classification score from negative samples’ feature $\mathbf{F}^{neg}$ as:
\begin{equation}
\begin{aligned}
\mathbf{S}^{neg} = \sigma(f_{cls}(\mathbf{F}^{neg})) \in \mathbb{R}^{\left|\mathcal{NG}\right|\times K}.
\label{Eq:bg loss}
\end{aligned}    
\end{equation}

\noindent\textbf{PBR Loss.} The PBR loss consists of the MIL loss $\mathcal{L}_{mil2}$ for positive bags and negative loss $\mathcal{L}_{neg}$ for negative samples, \ie,
\begin{equation}
\begin{aligned}
\mathcal{L}_{pbr} = \mathrm{\alpha_{2}} \mathcal{L}_{mil2}  +\mathrm{\alpha_{neg}} \mathcal{L}_{neg},
\label{Eq:PBR basic loss}
\end{aligned}
\end{equation}
\noindent where $\mathrm{\alpha_{2}}=0.25$ and  $\mathrm{\alpha_{neg}}=0.75$ are set here.

\noindent\textbf{1) MIL loss.} The MIL loss $\mathcal{L}_{mil2}$ in the PBR stage is defined as:
\begin{equation}
\begin{aligned}
\setlength\abovedisplayskip{0pt}
\setlength\belowdisplayskip{1pt}
\mathcal{L}_{mil2}  = \frac{1}{M}\sum\limits_{j=1}^{M} \left< \mathbf{c}^{\mathrm{T}}_j, \mathbf{\widehat{S}}^*_j \right> \cdot {\rm FL}(\mathbf{\widehat{S}}_j, \mathbf{c}_j),
\label{Eq:L_{bag}}
\end{aligned}
\end{equation}
\noindent where ${\rm FL}(\mathbf{\zeta}, \mathbf{\tau})$ is the focal loss~\cite{DBLP:retinanet_focalloss}, and $\gamma$ is set as 2 following~\cite{DBLP:retinanet_focalloss}. 
$\mathbf{\widehat{S}}^*_j$ represents the bag score of the last PBR iteration (we use the bag score in CBP for the first PBR iteration). ${\left< \mathbf{c}^{\mathrm{T}}_j, \mathbf{\widehat{S}}^*_j \right>}$ represents the inner product of the two vectors, meaning the predicted bag score of the previous stage or iteration on the GT category. It is used to weight the $\rm FL$ of each object for stable training.

\noindent\textbf{2) Negative loss.} Conventional MIL treats proposal boxes belonging to other categories as negative samples. To further suppress the backgrounds, we  introduce the negative loss ($\gamma$ is also set to 2 following $\rm FL$), \ie,

\begin{equation}
\begin{aligned}
\setlength\abovedisplayskip{1pt}
\setlength\belowdisplayskip{1pt}
\beta =\frac{1}{M}\sum\limits_{j=1}^{M} \left< \mathbf{c}^{\mathrm{T}}_j, \mathbf{\widehat{S}}^*_j \right>,
\label{Eq:neg_pbr1}
\end{aligned}
\end{equation}
\begin{equation}
\begin{aligned}
\setlength\abovedisplayskip{1pt}
\setlength\belowdisplayskip{1pt}
\mathcal{L}_{neg} = - \frac{1}{\left|\mathcal{NG}\right|}\sum\limits^{\left|\mathcal{NG}\right|}_{ng=1}\sum\limits_{k=1}^{K} \beta \cdot ([\mathbf{S}^{neg}_{ng}]_k)^{\gamma} \log(1-[\mathbf{S}^{neg}_{ng}]_k).
\label{Eq:neg_pbr2}
\end{aligned}
\end{equation}

\subsection{P2BNet++}
P2BNet++ is the first attempt in this paper  to transfer discrete optimization to continuous optimization in object perception. It processes discretely hand-crafted anchors, which are regressed to the boundaries of objects when constructing the proposal bags. This is achieved through a spatial self-distillation strategy in the cascade PBR stage.
\label{sec:p2b++}
\subsubsection{SPSD sampling}
Replacing basic sampling with SPSD Sampling in the last PBR stage yields P2BNet++. SPSD sampling means sampling with spatial position self-distillation (SPSD), which is proposed in SSD-Det~\cite{DBLP:SSD-Det}, which is a noisy-box supervised object detection network. 
It utilizes the statistical information of the whole dataset and conducts the regression module to self-distill the spatial position information. With the basic sampling, as shown in Fig.~\ref{fig:sample_cbp_pbr} (right), the proposals in a basic positive bag regress to the pseudo box of the previous stage. In this process, some proposals regress to the previous pseudo box, whereas others regress to the real ground truth (GT), which enhances the quality of proposal bags in the next stage. This is because many estimated boxes are high-quality in the whole dataset; the good samples teach the bad ones, which gives the network the ability to object-regression. It brings higher-IoU proposals, shown in Fig.~\ref{fig:ioudist}, guaranteeing good bag construction, which is better for MIL optimization and thus leads to better box selection.
\subsubsection{Loss function}
With the sampling strategy replaced, the MIL loss and negative loss in P2BNet++'s PBR stage are inherited from the original PBR stage. In addition, an extra L1 loss for box regression is introduced, as detailed in SSD-Det. The overall loss function of P2BNet++ consists of the CBP loss, multi-stage PBR loss, and box regression loss.

\subsection{Point-to-Mask Network}
\label{sec:p2mnet}

The P2MNet-MR framework consists of a Point-to-Mask Network (P2MNet) and Mask R-CNN (MR).
P2MNet predicts pseudo boxes and masks with point annotations to train the Mask R-CNN. Without fully annotated signal, the mask prediction faces challenges:

\noindent\textbf{Problem Statement:} When mask annotations are available, the task becomes a straightforward supervised 0-1 classification of the pixels within the target box. For instance, Mask R-CNN first performs object detection and then learns segmentation based on the detection results. However, in this task, only point annotations are provided. As there are no mask annotations to guide the box-to-mask prediction, it becomes an estimation problem. Using P2BNet (or P2BNet+) as the pseudo box predictor, P2MNet introduces an \textbf{O}bject-\textbf{a}ware \textbf{P}ixel \textbf{P}erception (OAPP) module to achieve precise pixel-level object representation. This is a further exploration of the discrete-to-continuous prediction approach following P2BNet++.

However, the bounding boxes obtained by P2BNet (or P2BNet++) are not guaranteed to be perfect. The prediction results are derived from candidate boxes sampled by discrete anchors, and manually set anchors cannot precisely fit target boundaries. This inevitably leads to predicted pseudo boxes that are either too large or too small. Smaller boxes may truncate the target and its corresponding mask, while larger ones introduce excessive background, causing the mask to misclassify background elements as foreground. As shown in Fig.~\ref{fig:Vis_analysis}, the predicted pseudo box (in red) truncates a waterbird's neck, resulting in a truncated mask due to the box constraint. In another example, the pseudo box does not closely fit the target, leading the predicted mask to include the bridge under the train as foreground.

\begin{figure*}[htb!]
    \centering
    \includegraphics[width=1.\linewidth]{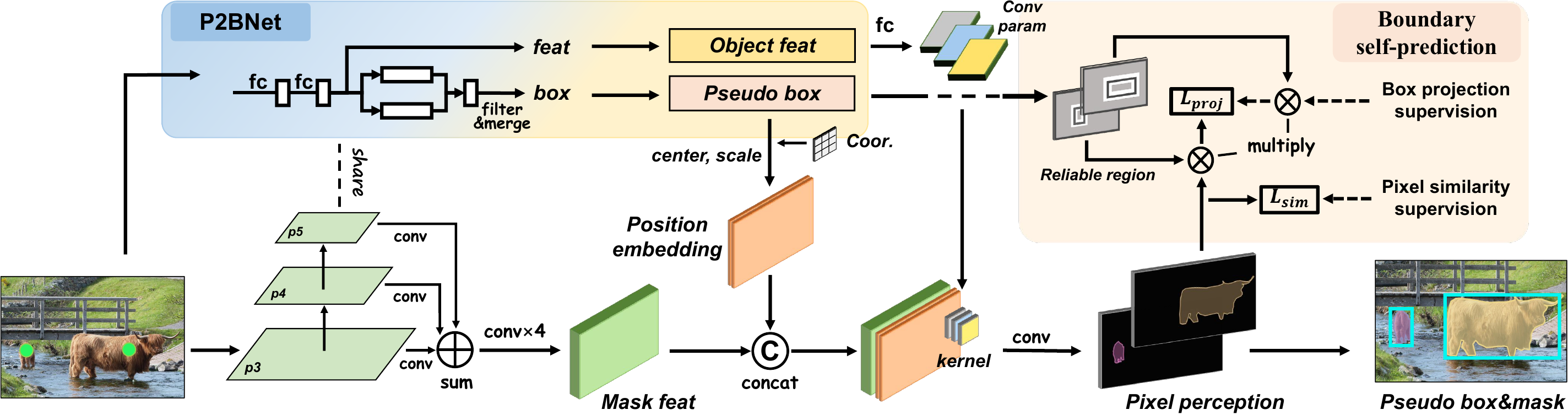}
    \vspace{-10pt}
     \caption{The framework of P2MNet. The P2BNet(or P2BNet++) serves as the pseudo box generator. The object feat from P2BNet predicts the conv parameter and the pseudo box generates the position embedding. The position embedding is combined with the mask feat, and the kernel with the predicted conv parameter is conducted on the feature map to estimate the pixel-level perception of each object. The pixel perception mask is supervised by box projection and pixel similarity. Then, the minimum circumscribed rectangle of the pixel perception mask is the refined pseudo box.}
     \vspace{-12pt}
    \label{fig:p2mnet framework}
\end{figure*}

\subsubsection{Overall Framework}
The architecture of P2MNet is shown in Fig.~\ref{fig:p2mnet framework},
where P2BNet, serving as a component of P2MNet, predicts the pseudo box as weak supervision for mask estimation and object feat to predict dynamic convolution kernel parameter.  
Motivated by the box-supervised instance segmentation method BoxInst~\cite{DBLP:condinst, DBLP:Boxinst}, the dynamic convolution kernel of each object acts on the mask feature to obtain a respective object mask. It is supervised by the box projection loss with pseudo box and pixel similarity loss with image color.
%
The background regions outside the bounding box are predicted as negative examples due to the box projection loss constraint. Areas within the bounding box that are connected with the exterior background must be classified similarly to those outside, due to the pixel similarity loss constraint, resulting in their classification as negative instances as well. Through this mechanism, different background regions inside the box are progressively excluded, and the remaining part represents the foreground mask of the target.

The above assumes that the box is the ground truth box. However, the pseudo boxes predicted by P2BNet have deviations and cannot closely fit the target boundaries, leading to projection constraint losses from the pseudo boxes that either truncate the target or include excessive background. The BSP designed in the paper aims to address this issue. Through surface analysis, the image color similarity constraint is found to be related only to the image itself, rather than the quality of the annotation box, making the color similarity loss in mask prediction reasonable. Since P2BNet's pseudo box annotations have certain deviations, we can rely on their estimated approximate scale and shape, but not on the precise boundary contours. BSP is designed to relax the projection constraint at the target boundary, allowing for self-correction of the boundary based on the network’s learned statistical probability within the approximate pseudo-box range, thus refining the target contour.

The overall loss function of P2MNet is the summation of P2BNet loss $\mathcal{L}_{p2b}$, box projection loss $\mathcal{L}_{proj}$ and pixel similarity loss $\mathcal{L}_{sim}$, \ie,

\begin{equation}
\begin{aligned}
\mathcal{L}_{p2m}=  \mathrm{\alpha_{3}}\mathcal{L}_{p2b} +  \mathrm{\alpha_{4}}\mathcal{L}_{proj}+ \mathrm{\alpha_{5}}\mathcal{L}_{sim}, \\
\label{Eq:P2M_loss_all}
\end{aligned}
\end{equation}

\noindent where loss weights $\mathrm{\alpha_{3}}$, $\mathrm{\alpha_{4}}$ and $\mathrm{\alpha_{5}}$ are set as 1 here. 

\subsubsection{Object-aware Pixel Perception}
\label{sec:oapp}

\noindent\textbf{Network Structure.}
As shown in Fig.~\ref{fig:p2mnet framework}, each feature of FPN (P3, P4, P5) passes a 3 $\times$ 3 convolution to create the refined feature map (from 256 channels to 128 channels). Then, the feature maps of P4 and P5 upsample to the P3 level shape, and the values of the three levels' features are added.
Similar to ~\cite{DBLP:condinst,DBLP:Boxinst}, for the backbone features, a mask branch is connected to the FPN P3 level to obtain $\mathbf{F}_{mask} \in \mathbb{R}^{D\times H\times W}$ whose output resolution is 1/8 of the input image resolution. The mask branch consists of four 3 $\times$ 3 convolutions with 128 channels before the last layer. The last layer of the mask branch reduces the number of channels from 128 to 16.

In the last stage of P2BNet, the top-$k$ proposal boxes of each object with the highest proposal scores are weighted to obtain the pseudo boxes $\mathbf{b}_j \in \mathbb{R}^{4} $. 
Also, the top-$k$ proposal features are weighted by the proposal scores and summed to obtain the object feat $\mathbf{\hat{F}}_j$. 
Each object's pseudo boxes can provide the position embedding, which helps predict respective masks for different objects. 
We calculate the offset of the x-axis and y-axis from every grid center of the feature map $\mathbf{F}_{mask}$ to the center $(\mathbf{b}_{jx},\mathbf{b}_{jy})$ of the pseudo box $\mathbf{b}_j$.
Then, in order to normalize the offsets by object scale, the coordinate distance maps are divided by the adjust factor (64, 128, 256, and 512), respectively, for pseudo boxes assigned to corresponding FPN levels. That serves as the position embedding, defined as $\mathbf{PE} \in \mathbb{R}^{M\times 2\times H\times W}$. $M$ is the number of objects, and the dimension of `2' means the x-axis and y-axis 
offsets. Then, the feature map $\mathbf{\hat{F}}_{mask}$ is obtained by:
\begin{equation}
\mathbf{\hat{F}}_{mask}= \mathbf{F}_{mask} \otimes \mathbf{PE}  \in \mathbb{R}^{M\times (D+2)\times H\times W,} 
\end{equation}
\noindent where $\otimes$ is the concat operation, and mask feature $\mathbf{F}_{mask}$ is broadcast into eight copies. Here, the dimension D is 16 as mentioned above.

Then a linear layer $f_{par}$ is adopted to obtain the convolution parameters $\mathbf{\theta}_j$, \ie,
\begin{equation}
    \mathbf{\theta}_j = f_{par}(\mathbf{\hat{F}}_j) \in \mathbb{R}^G,
    \label{param}
\end{equation}
\noindent where $G$ is the total number of the dynamic convolution kernel weights and biases. 

The parameters $\mathbf{\theta}_j$ can be split into three convolution kernel weights and biases. The total number G is 233 (\#weights = (16 + 2) $\times$ 8(conv1) + 8 $\times$ 8(conv2) + 8$\times$ 1(conv3) and \#biases = 8(conv1) + 8(conv2) + 1(conv3)). $\mathbf{\hat{F}}_{mask}$ passing different convolution kernels can generate different object score maps for different objects. After the dynamic convolution and sigmoid activation function, the object score map $\mathbf{S}^{map} \in \mathbb{R}^{M\times H\times W}$ is obtained. Object masks are defined as:
\begin{equation}
    \mathbf{Mk}=\mathbf{S}^{map} > \chi_1 \in \{0,1\}^{M\times H\times W,}
\end{equation}
where $\chi_1$ = 0.5. The pseudo box is updated by the minimum external rectangle of each object mask.


\noindent\textbf{Loss Function.} The loss function contains box projection loss and pixel similarity loss:

\noindent\textbf{1) Box projection loss} uses a bounding box to supervise mask prediction. 
We treat the pseudo box generated from P2BNet (or P2BNet++) as a mask $\mathbf{Mk}^b \in \{0,1\}^{M\times H\times W} $,  by assigning 1 to pixels inside the box and 0 to pixels outside the box.
Then, we have $Proj_x(\mathbf{Mk}^b)$ and $Proj_y(\mathbf{Mk}^b)$, where $Proj_x: \mathbb{R}^{M \times H \times W} \rightarrow \mathbb{R}^{M \times W}$ and $Proj_y: \mathbb{R}^{M \times H \times W} \rightarrow \mathbb{R}^{M \times H}$ denote the projection of the mask onto the x-axis and y-axis. The projection operation can be implemented by a max operation along with each axis. Formally, we define $Proj_{x/y}(\mathbf{Mk}^b)=Max_{y/x}(\mathbf{Mk}^b)$. 
Also, the predicted object mask can conduct the same operation, and the projection can be defined as $Proj_{x/y}(\mathbf{S}^{map})=Max_{y/x}(\mathbf{S}^{map})$.
We then compute the loss between the projection of the (pseudo) GT box and predicted object masks. The projection loss term is defined in Eq.~\ref{Eq:proj(a,b)}:
\begin{equation}
\begin{aligned}
    \mathcal{L}_{proj}&=\frac{1}{M}\sum\limits_{j=1}^{M} Proj(\mathbf{S}^{map}_j,\mathbf{Mk}^b_j)\\
    &=\frac{1}{M}\sum\limits_{j=1}^{M} \big\{\mathrm{DC}(Proj_x(\mathbf{S}^{map}_j), Proj_x(\mathbf{Mk}^b_j))\\
    &+\mathrm{DC}(Proj_y(\mathbf{S}^{map}_j), Proj_y(\mathbf{Mk}^b_j))\big\}\\
    &=\frac{1}{M}\sum\limits_{j=1}^{M} \big\{\mathrm{DC}(Max_y(\mathbf{S}^{map}_j), Max_y(\mathbf{Mk}^b_j))\\ 
    &+\mathrm{DC}(Max_x(\mathbf{S}^{map}_j), Max_x(\mathbf{Mk}^b_j))\big\},\\
    \label{Eq:proj(a,b)}
\end{aligned}
\end{equation}
where $j$ is the $j$-th object in the image. The $\mathrm{DC}(\mathbf{\zeta}, \mathbf{\tau})$ is the Dice loss:

\begin{equation}
\begin{aligned}
\mathrm{DC}(\mathbf{\zeta}, \mathbf{\tau})=1-\frac{2\mid \mathbf{\zeta} \cap \mathbf{\tau} \mid}{\mid \mathbf{\zeta} \mid + \mid \mathbf{\tau} \mid},
\end{aligned}
\end{equation}
\noindent  To some extent, this is also a multiple instance learning paradigm. Hence, it is a soft loss style. Only the max-score grid in every row or column participates in loss calculation every time. 
However, owing to the inaccurateness of the pseudo box mentioned in Fig.~\ref{fig:Vis_analysis} and the problems resulting therefrom, we design the BSP for compensation, described in Sec.~\ref{sec:BSP}.

\noindent\textbf{2) Pixel similarity loss} is used to help predict more accurate boundaries and distinguish foreground and background. It leverages the prior that if they have similar colors in the raw images, the proximal pixels are likely to be in the same class, \ie, the foreground or background.
First, we calculate the color similarity of the given images in LAB color space as $\iota \in \mathbb{R}^{8 \times H\times W}$, and its details are described in BoxInst.
Here, $H\times W$ is the number of pixel points in the image, and the dimension `8' denotes the neighborhood points of 8 directions (left-up, up, right-up, right, right-down, down, left-down, and left) of each pixel-point. $\iota$ ranges from 0 to 1 where `1' means very similar (same) and `0' means very different. Second, we compute the pixel similarity loss for similar pairwise and discard those with agnostic labels (whose similarity is lower than a threshold $\chi_2$). The pixel similarity loss is defined in Eq.~\ref{Eq:similar_loss}.
\begin{equation}
\begin{aligned}
    \mathcal{L}_{sim}&=\frac{1}{M}\sum\limits_{j=1}^{M} Sim(S^{map}_j, \mathbf{Mk}^b_j) \\
     & =-\frac{1}{M}\frac{1}{N_j}\sum\limits_{j=1}^{M}\sum\limits_{\omega \in \Omega(\mathbf{Mk}_j^b)} \mathrm{1}_{\{\iota_\omega \ge \chi_2\}}logP_{sim}(\omega),
\end{aligned}
    \label{Eq:similar_loss}
\end{equation}

\noindent where $\Omega(\mathbf{Mk}_j^b)$ \label{sim loss}denotes the set of the pairwise in which at least one pixel inside the positive region of $\mathbf{Mk}^b_j$, and $N_j$ is the length of $\Omega_j$. $\iota_\omega$ denotes the pairwise. $P_{sim}(\omega)$ means the probability of similarity, \ie,
\begin{equation}
    P_{sim}(\omega)=(\mathbf{S}^{map}_{\omega_{1}}) (\mathbf{S}^{map}_{\omega_{2}})+(1-\mathbf{S}^{map}_{\omega_{1}}) (1-\mathbf{S}^{map}_{\omega_{2}}),
\end{equation}

\noindent where $\omega_1$, $\omega_2$ means the pairwise pixel points, and ${S}^{map}_{\omega_{1}}$ and ${S}^{map}_{\omega_{2}}$ are their predicted scores on $S^{map}_j$.


\subsubsection{Boundary Self-Prediction}
\label{sec:BSP}
To address problems of truncated and overinclusive predictions caused by the imprecisely estimated bounding box, we design a \textbf{B}oundary \textbf{S}elf-\textbf{P}rediction (BSP) to obtain a more accurate object boundary. 
Because the uncertain area is often at the object boundary, the project loss’s boundary region is unreliable. Hence, we calculate the reliability map of the object with the bounding box $\mathbf{b}_j=(\mathbf{b}_{jx},\mathbf{b}_{jy},\mathbf{b}_{jw},\mathbf{b}_{jh})$. We define the uncertainty ratio as $\eta$ and the reliability map as $\mathbf{A}_j^{proj}$, and $\mathbf{A}_j^{sim}\in \mathbb{R}^{W \times H}$ is defined in Algorithm~\ref{Alg: RM}. Then the reliability map $\mathbf{A}_j^{proj} \in \mathbb{R}^{M \times W \times H}$ for all objects in the image is multiplied by the object score map $\mathbf{S}^{map}$ and $\mathbf{Mk}^b$ to calculate the box projection loss:
\begin{equation}
\begin{aligned}
    \mathcal{L}_{proj}&=\frac{1}{M}\sum\limits_{j=1}^{M} Proj(\mathbf{S}^{map}_j\odot \mathbf{A}^{proj}_j,\mathbf{Mk}^b_j\odot \mathbf{A}^{proj}_j)
\end{aligned}
\end{equation}
To expand the region of the pixel similarity loss, the $\mathbf{A}^{sim} $ is used to replace the $\mathbf{Mk}^b$,
the loss function is defined as:

\begin{equation}
\begin{aligned}
    \mathcal{L}_{sim}&=\frac{1}{M}\sum\limits_{j=1}^{M} Sim(\mathbf{S}^{map}_j, \mathbf{A}^{sim}_j).
\end{aligned}
\end{equation}
\noindent $Proj(\cdot,\cdot)$ and $Sim(\cdot,\cdot)$ are defined in Eq.~\ref{Eq:proj(a,b)}, \ref{Eq:similar_loss}.

\begin{algorithm}[tb!]
{
\small
\caption{Reliability Map Calculation}
\label{Alg: RM}
\textbf{Input:}
pseudo box $\mathbf{b}_j=(\mathbf{b}_{jx},\mathbf{b}_{jy},\mathbf{b}_{jw},\mathbf{b}_{jh})$, uncertainty ratio $\eta$ and object mask shape $W,H$.\\
\textbf{Output:} Reliability Maps $\mathbf{A}_j^{proj}$ and $\mathbf{A}_j^{sim}$.\\
\textbf{Note:} $\mathbf{\vec{m_1}},\mathbf{\vec{m_2}}$ are two W$\times$H matrix (initialized by 0).\\
\vspace{-10pt}
\begin{algorithmic}[1]
\STATE  $\mathbf{\vec{m_1}},\mathbf{\vec{m_2}} \leftarrow  \boldsymbol{O}, \boldsymbol{O}$;
\STATE \% the left, right, bottom and up of the given box
\STATE $\mathbf{b}_{jl}, \mathbf{b}_{jr} = \mathbf{b}_{jx} - \mathbf{b}_{jw}/ 2, \mathbf{b}_{jx} + \mathbf{b}_{jw}/ 2$;
\STATE $\mathbf{b}_{jb}, \mathbf{b}_{ju} = \mathbf{b}_{jy} - \mathbf{b}_{jh}/ 2, \mathbf{b}_{jy} + \mathbf{b}_{jh}/ 2$;
\STATE  $\vec{m_1}[\mathbf{b}_{jl}+\mathbf{\eta}*\mathbf{b}_{jw} : \mathbf{b}_{jr}-\mathbf{\eta}*\mathbf{b}_{jw}, \mathbf{b}_{jb}+\mathbf{\eta}*\mathbf{b}_{jh} : \mathbf{b}_{ju}-\mathbf{\eta}*\mathbf{b}_{jh}]=1$
\STATE  $\vec{m_2}[\mathbf{b}_{jl}-\mathbf{\eta}*\mathbf{b}_{jw} : \mathbf{b}_{jr}+\mathbf{\eta}*\mathbf{b}_{jw}, \mathbf{b}_{jb}-\mathbf{\eta}*\mathbf{b}_{jh} : \mathbf{b}_{ju}+\mathbf{\eta}*\mathbf{b}_{jh}]=1$
\STATE  \% $\mathbf{A}_j^{proj}$ is the uncertainty region around edge
\STATE $\mathbf{A}_j^{proj}\leftarrow(\mathbf{\vec{m_1}}+\mathbf{\vec{m_2}})\neq1$
\STATE $\mathbf{A}_j^{sim} \leftarrow \mathbf{\vec{m_2}}$ 

\end{algorithmic}
}
\end{algorithm}

\section{Experiments}
\label{sec:experi}
\subsection{Experiment Settings}
\textbf{Datasets.} For experimental comparisons, we use four publicly
available datasets: MS COCO~\cite{coco}, Pascal VOC~\cite{DBLP:VOC}, SBD~\cite{DBLP:VOC} and Cityscapes~\cite{cityscape}. \textbf{MS COCO} has 80 different categories and two versions. COCO-14 has 80K training and 40K validation images, whereas COCO-17 has 118K training and 5K validation images. Considering that the GT on the test set is not released, we train our model on the training set and evaluate it on the validation set.
\textbf{Pascal VOC} contains 20 object categories. In VOC 2007, the \texttt{trainval} set has 5,011 (2,501\,+\,2,510) images for model training and the \texttt{test} set contains 4,951 images for evaluation. In VOC 2012, 11,540 (5,717\,+\,5,823) images are divided into the \texttt{trainval} set for training and 10,991 images belong to the \texttt{test} set. \textbf{SBD} is an enhanced dataset of VOC2012 for the instance segmentation task. It has 20 categories and contains 10582 images in \texttt{train} set and 1449 \texttt{val} images for evaluation. \textbf{Cityscapes} is a street view dataset with eight categories for object detection and instance segmentation. 2975 images are for train and 500 images are for evaluation.

\noindent\textbf{Evaluate metrics.}
We report AP (the averaged over IoU thresholds in $[0.5:0.05:0.95]$), AP$_{50}$, AP$_{75}$ and AP on small, medium and large on COCO. The mIoU$_{box/mask}$ is calculated by the mean IoU between the predicted pseudo boxes or masks and their corresponding GT of all objects in the training set to show the quality of estimated pseudo labels. It can directly evaluate P2BNet's ability to transform the annotated points into accurate pseudo boxes. For VOC detection, we report the AP$_{50}$, and for SBD segmentation, we report AP$_{25, 50, 75}$. For Cityscapes, we report AP and AP$_{50}$ for detection and segmentation.

\noindent\textbf{Implementation details.} \label{imple-detail} Our codes of P2BNet-FR (and P2MNet-FR) are based on MMDetection~\cite{mmdetection}. 
The stochastic gradient descent (SGD~\cite{DBLP:series/lncs/Bottou12}) algorithm is used to optimize in 1$\times$ training schedule. The learning rate is set to 0.02 and decays at the 8th and 11th epochs. 
We use multi-scale (480-1200) as the short side to resize the image during training and single-scale (1200) during inference. The mini-batch has 16 images; all models are trained with 8 GPUs for the COCO dataset. 
Loss weights are fixed during training and set as $\alpha_{1}=0.25$ in CBP, $\alpha_{2}=0.25$ and $\alpha_{neg}=0.75$ in PBR, following the focal loss~\cite{DBLP:retinanet_focalloss}.  In P2MNet, each object's score in P2BNet++ serves as the dynamic weight to multiply at $\alpha_{4}$ and $\alpha_{5}$.

We choose the classic Faster R-CNN FPN~\cite{FasterRCNN,DBLP:FPN} and Mask R-CNN (with ResNet-50~\cite{DBLP:resnet}) as the detector and segmentor with the default setting. The single-scale (800) images are used during training and inference. 
In CBP sampling, $s\in \{4, 8, 16, 32, 64, 128\} \cdot \delta$, where $\delta$ is a factor for dynamic adjustment according to the dataset. $\delta=min(W,H)/100$, where $W$ and $H$ are the image's width and height. $v\in \{1/3, 1/2, 2/3, 1, 3/2, 2, 3\}$ is used as the fixed setting. We deem $(v\cdot s)$ and $(v/s)$ entireties in the PBR sampling. We set $(v\cdot s) \in \{0.7, 0.8, 1, 1.2, 1.3\}$, $(v/s) \in \{0.7, 0.8, 1, 1.2, 1.3\}$, and thus, we have $5\times5=25$ options. $(o_x,o_y) \in \{(0,0), (1,0), (0,1), (-1,0), (-1,-1)\}$ is used to jitter the center position. 
In P2MNet,  the region rate $\mathbf{\eta}$ of BSP is set as 0.1 by default.

\begin{table*}[tb!]
    \setlength{\tabcolsep}{9pt}
    \centering
    \resizebox{1.0\textwidth}{!}{
    \begin{tabular}{l|c|c|ccccc|ccccc}
    \specialrule{0.13em}{0pt}{1pt}  
     \multirow{2}{*}{Method} & \multirow{2}{*}{Backbone} & \multirow{2}{*}{Proposal}  & \multicolumn{5}{c|}{COCO-14} & \multicolumn{5}{c}{COCO-17}  \\
    &&& AP & AP$_{50}$ & AP$^s$&AP$^m$ &AP$^l$& AP & AP$_{50}$&AP$^s$&AP$^m$ &AP$^l$ \\
    \specialrule{0.08em}{0pt}{1pt}
    \multicolumn{5}{l}{\textbf{\emph{Box-supervised detectors}}} \\
    \specialrule{0.08em}{0pt}{0pt}
     Fast R-CNN~\cite{fastrcnn}  & VGG-16 &SS  & 18.9 & 38.6&-&-&- &19.3&39.3&-&-&-\\
    Faster R-CNN~\cite{FasterRCNN} & VGG-16 & RPN  & 21.2 &41.5&-&-&- &21.5 &42.1&-&-&- \\
    FPN~\cite{FasterRCNN,DBLP:FPN} &R-50&RPN &35.5&56.7& 19.6& 38.2 &44.1 &37.4&58.1&21.2&41.0&48.1\\
    RetinaNet~\cite{DBLP:retinanet_focalloss}  &R-50&-&34.3&53.3&-&-&-&36.5&55.4&20.4&40.3&48.1 \\
    Reppoint~\cite{DBLP:reppoint}  &R-50&-&-&-&-&-&-&37.0&56.7&20.4&41.0&49.0 \\
    Sparse R-CNN$^\ast$~\cite{DBLP:sparsercnn} &R-50&PP&-&-&-&-&-&45.0 &64.1 & 28.0 &47.6& 59.5  \\  
    Swin-Transformer MR$^\ast$~\cite{DBLP:swin-transformer} &Sw-S&RPN&-&-&-&-&-&48.2& 69.8 & 32.1& 51.8& 62.7 \\
    \specialrule{0.08em}{0pt}{1pt}
    \multicolumn{5}{l}{\textbf{\emph{Image-supervised detectors}}} \\
    \specialrule{0.08em}{0pt}{0pt}
    OICR+Fast~\cite{DBLP:oicr,fastrcnn} &  VGG-16 &SS&7.7&17.4&-&-&- &-&-&-&-&-\\
     OICR+REG~\cite{DBLP:oicr,fastrcnn} &  VGG-16 &SS&-&-&-&-&-&9.8 &20.8& 1.4& 9.2& 17.7
\\
     OICR+REG~\cite{DBLP:oicr,fastrcnn}+W2N~\cite{DBLP:W2N} &  V16+R50 &SS&-&-&-&-&-&15.3 &30.0 & 4.9& 18.5& 24.6\\
     PCL~\cite{DBLP:pcl} & VGG-16&SS& 8.5&19.4&-&-&-&-&-&-&-&-\\   
     MEFF+Fast~\cite{MEFF,fastrcnn}  &  VGG-16&SS& 8.9&19.3&-&-&- &-&-&-&-&-\\
     C-MIDN~\cite{C-MIDN} &VGG-16& SS& 9.6&21.4&-&-&-&-&-&-&-&-\\
    WSOD2~\cite{WSOD2} & VGG-16&SS& 10.8&22.7&-&-&-&-&-&-&-&-\\
    UFO~\cite{UFO2} &VGG-16& MCG  &10.8 &23.1 &-&-&-&-&-&-&-&-\\
    GradingNet-C-MIL~\cite{DBLP:guidingnet} & VGG-16 & SS & 11.6 & 25.0 &-&-&-&-&-&-&-&-\\
     ICMWSD~\cite{ICMWSD} & VGG-16 & MCG  & 11.4 &24.3&3.6&12.2&17.6&12.4&25.8&3.9&13.8&19.9\\
       ICMWSD~\cite{ICMWSD}     &   R-50 & MCG   & 12.6 &26.1 &3.7&13.3&19.9&-   &-  &-&-&-  \\
     CASD~\cite{CASD} &  VGG-16 & SS  &12.8 & 26.4  &-   &-&-&-&-&-&-&-  \\
     CASD~\cite{CASD}    &     R-50 & SS   & 13.9&27.8 &-   &- &-&10.5& 24.1& 2.7& 12.2& 18.3    \\ 
     CASD~\cite{CASD}+W2N~\cite{DBLP:W2N}    &     R-50 & SS   & -&- &-   &-&-&15.9 &33.3 & 5.6 &18.4 &27.2    \\ 
     OD-WSCL~\cite{od-wscl} &VGG-16&SS+MCG&13.7& 27.7& 4.4& 14.5& 21.2&13.6 &27.4& 4.9 &15.5 &21.6 \\
     OD-WSCL~\cite{od-wscl} &R-50&SS+MCG&13.9 &29.1 &4.9& 16.8& 22.3&13.8& 27.8 &5.7& 17.7 &23.8  \\ 
    \specialrule{0.08em}{0pt}{1pt}
    \multicolumn{5}{l}{\textbf{\emph{Point-supervised detectors}}} \\
    \specialrule{0.08em}{0pt}{0pt}
     Click~\cite{Click}   & AlexNet&SS  & - &18.4 &-&-&-&-&-&-&-&- \\
     UFO$^2$~\cite{UFO2}      &VGG-16& MCG  &12.4 &27.0&-&-&-&-&-&-&-&-\\
    UFO$^{2{\dagger}}$~\cite{UFO2}     &VGG-16& MCG    &12.8 &26.6&4.0&13.3&20.1 &13.2 &27.2&4.3&14.6&21.4\\ 
    UFO$^{2{\ddagger}}$~\cite{UFO2}  &VGG-16& MCG  &12.7& 26.5 &3.9&13.5&20.0&13.5 &27.9 &4.4&15.0&22.0\\
    UFO$^{2{\ddagger}}$~\cite{UFO2}   & R-50 & MCG &12.6&27.6&4.5&12.5&19.6&13.2 &28.9 &5.2&14.2&21.2\\    
    \rowcolor{Gray}
     P2BNet-FR (Ours)   & R-50 & Free   &19.4&43.5 &9.3&21.2&25.6& 22.1 & 47.3 &11.5&24.8&30.4\\
     \rowcolor{Gray}
      P2BNet++-FR (Ours)   & R-50 & Free &21.7&  {45.7}&  {10.1}&  {23.3}&  {29.3}&  {23.1}&  {47.4}&  {11.0}&   {25.4} &  {32.9} \\
\rowcolor{Gray}
        {P2MNet-MR (Ours)}   &   {R-50} &  {Free}   &  {24.3} &  {46.9}&  {11.6}&  {26.1}&  {32.3}&  {25.6}&   {48.4}&   {13.1} &  {28.7} &  {35.3} \\
      \rowcolor{Gray}
         {P2MNet-FR (Ours) }  &  {R-50} &  {Free}   &  {24.5} &  {47.2}&  {11.6}&  {26.3}&  {32.6}&  {25.9}&  {49.0}&  {12.9}&  {29.0}&  {36.0} \\
    \rowcolor{Gray}
     {P2MNet-SR}$^\ast$(Ours)&  {R-50}&  {Free}&  {28.1}&  {50.2}&  {14.1}&  {30.0}&  {38.1}&  {29.6}&  {52.5}&  {16.9}&  {32.7}&  {42.3}\\
    \rowcolor{Gray}
        {P2MNet-SwinMR$^\ast$ (Ours)}   &   {R-50+Sw-S} &   {Free}   &  {\textbf{28.8}}&  {\textbf{54.3}}&  {\textbf{15.2}}&  {\textbf{30.9}}&  {\textbf{38.7}}&  {\textbf{30.3}}&  {\textbf{55.9}}&  {\textbf{17.9}}&  {\textbf{33.3}}&  {\textbf{41.7}} \\
    \specialrule{0.13em}{0pt}{0pt}
    \end{tabular}
    }
    \caption{The performance comparison of box-, image-, and point-supervised detectors on COCO val dataset. The performance of P2BNet-FR, UFO$^2$, and the box-supervised detector is tested on a single-scale dataset.
    $^{\dagger}$ means we reproduce the performance with the original setting. $^{\ddagger}$ means we re-implement UFO$^2$ with our QC point annotation. $^\ast$ means the multi-scale training on the retrain detector. 
    SS is selective search~\cite{DBLP:selectivesearch}, PP means proposal box defined in \cite{DBLP:sparsercnn}, and Free represents the \otsp-free based method. Sw-* is the Swin-Transformer backbone. FR, MR, SR and SwinMR are the advanced detectors: Faster R-CNN~\cite{FasterRCNN}, Mask R-CNN~\cite{DBLP:MASK-RCNN}, Sparse R-CNN~\cite{DBLP:sparsercnn} and Swin-Transformer Mask R-CNN~\cite{DBLP:swin-transformer}.}
    \vspace{-10pt}
    \label{tab:coco-tmp}
\end{table*}

\noindent\textbf{Quasi-Center point annotation.} We propose a quasi-center (QC) point annotation that is friendly for object detection tasks with a low cost. In practical scenarios, we ask annotators to annotate the object in the non-high limit center region with a loose rule. Since datasets in the experiment are already annotated with bounding boxes or masks, it is reasonable that the manually annotated points follow the Gaussian distribution in the central region.
We utilize the Rectified Gaussian ($RG$) Distribution defined in \cite{CPR} with central ellipse constraints.
For a bounding box of $b=(b_x,b_y,b_w,b_h)$, its central ellipse can be defined as $Ellipse(\kappa)$, using $(b_x,b_y)$ as the ellipse center and $(\kappa \cdot b_w, \kappa \cdot b_h)$ as the two axes of the ellipse. 
If the object's GT mask overlaps with the central ellipse $Ellipse(\kappa)$, $\mathcal{V}$ is used to denote the point set of the intersection. If there is no intersecting area, $\mathcal{V}$ represents the point set of the entire GT mask. 
When generated from bounding box annotations, the boxes are treated as masks. $RG$ is defined as:
\begin{equation}
RG(p;\mu,\sigma , \kappa)=\left\{
\begin{array}{rcl}
& \frac{Gauss(vs;\mu,\sigma)}{\int_\mathcal{V} Gauss(\mathbf{p};\mu,\sigma)d\mathbf{p}}, & {\mathbf{p} \in \mathcal{V}}\\
& 0,  & {\mathbf{p} \notin \mathcal{V}}
\end{array} \right.
\end{equation}
\noindent where $\mu$ and $\sigma$ are mean and standard deviation of $RG$. $\kappa$ decides the $Ellipse(\kappa)$. In this paper, $RG(p;0, \frac{1}{4},\frac{1}{4})$ is chosen. 

\begin{figure*}[tb!]
        \centering
        \includegraphics[width=1.\linewidth]{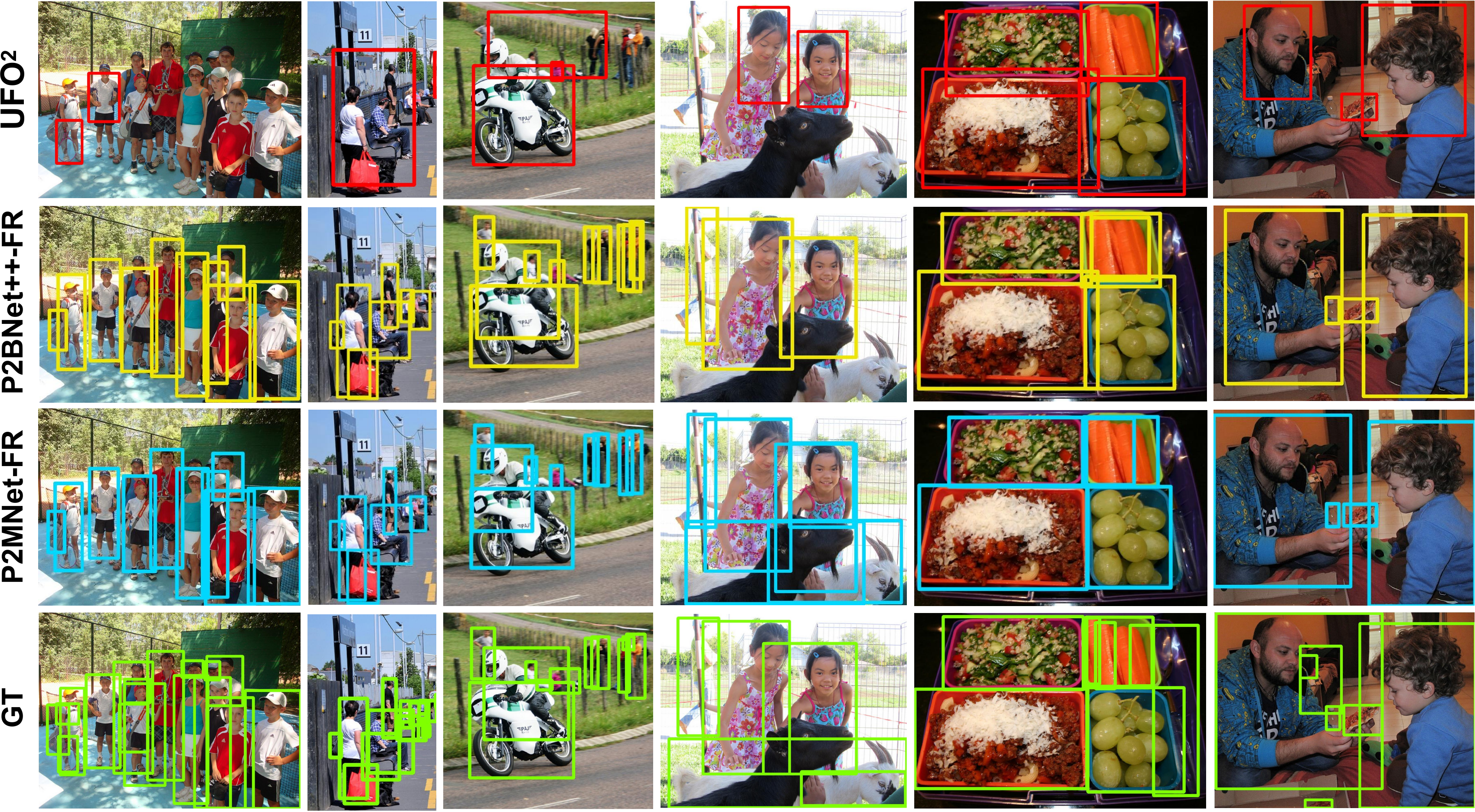}
    \caption{Visualization of the detection results of P2MNet-FR, P2BNet++-FR and UFO$^2$. Compared to UFO$^2$, our P2MNet-FR framework can distinguish dense objects and perform well in complex scenes. Compared to P2BNet++-FR, our P2MNet-FR framework misses a few objects (column 4) and predicts tighter bounding boxes (columns 5 and 6). (Best viewed in color.)}
    \vspace{-0.3cm}
    \label{fig:Vis}
\end{figure*}
\begin{table}[tb!]
    \centering
    \resizebox{1.\linewidth}{!}{
    \begin{tabular}{l|ccccc}
    \specialrule{0.13em}{0pt}{0pt}
         Method&\makebox[0.03\textwidth][c]{Sup.}&\makebox[0.04\textwidth][c]{BB.}&\makebox[0.03\textwidth][c]{Proposal}&\makebox[0.06\textwidth][c]{VOC2007}&\makebox[0.055\textwidth][c]{VOC2012}\\
             \specialrule{0.08em}{0pt}{0pt}
        \color{gray} {FPN~\cite{FasterRCNN,DBLP:FPN}} &\color{gray} {$\mathcal{B}$}&\color{gray} {R-50}&\color{gray} {SS}&\color{gray} {77.2}&\color{gray} {75.3}\\
        \color{gray} {CASD~\cite{CASD} }&\color{gray} {$\mathcal{I}$}&\color{gray} {VGG-16}&\color{gray} {SS}&\color{gray} {56.8}&\color{gray} {53.6}\\
        \color{gray} {OD-WSCL~\cite{od-wscl}} & \color{gray} {$\mathcal{I}$}&\color{gray} {VGG-16}&\color{gray} {MCG}&\color{gray} {13.6}&\color{gray} {56.2}\\
        Click~\cite{Click}&$\mathcal{P}$&AlexNet&SS&45.9&-\\
        UFO$^2$~\cite{UFO2}&$\mathcal{P}$&VGG-16&MCG&57.6&41.0\\
        UFO$^{2{\ddagger}}$~\cite{UFO2}&$\mathcal{P}$&R-50&MCG&57.6&38.6 \\
         \specialrule{0.08em}{0pt}{0pt}  
\multicolumn{5}{l}{\textbf{\emph{Ours}}} \\
\specialrule{0.08em}{0pt}{0pt} 
\rowcolor{Gray}
        P2BNet-FR&$\mathcal{P}$&R-50&Free&60.4&60.8\\
        \rowcolor{Gray}
        P2BNet++-FR&$\mathcal{P}$&R-50&Free& 66.2 & 62.3\\
        \rowcolor{Gray}
        P2MNet-FR&$\mathcal{P}$&R-50&Free& \textbf{68.2} &\textbf{66.7}\\
        \specialrule{0.13em}{0pt}{0pt}
    \end{tabular}}
    \caption{Object detection performance on VOC 2007 and 2012 $test$ sets. The evaluation metric is $\rm AP_{50}$ and the retrained detector of our method is trained on a single-scale dataset. Here, `B' (Box-level) can be viewed as 4 points for supervision, we named it `4-points', `I' (Image-level) can be named as `0-point', `P' (Point-level) can be named as `1-point'.}
    \vspace{-15pt}
    \label{tab:VOC_DET}
\end{table}

\subsection{Object Detection Results}
\label{sec:psod}
\subsubsection{Experiments on COCO}

Unless explicitly stated, our P2MNet-FR framework uses P2MNet and Faster R-CNN as the default components. We compare the P2MNet-FR and P2BNet++-FR with the existing PSOD methods while choosing the state-of-the-art UFO$^2$~\cite{UFO2} framework as the baseline for comprehensive comparisons.
Furthermore, we compare the advantages of our PSOD methods’ performance to state-of-the-art WSOD methods. To establish the performance upper bound, we also compare with the box-supervised object detectors.

\noindent\textbf{Comparison with the PSOD methods.}
As shown in Table~\ref{tab:coco-tmp}, we compare the existing PSOD methods Click~\cite{Click} and UFO$^2$~\cite{UFO2} on COCO. Both Click, and UFO$^2$ utilize \otsp-based methods (SS~\cite{DBLP:selectivesearch} or MCG~\cite{DBLP:MCG}) to generate proposal boxes. We use the public code to retrain UFO$^2$ with our QC point annotation to ensure a fair comparison. In addition, the previous methods are mainly based on VGG-16~\cite{DBLP:VGG} 
or AlexNet~\cite{DBLP:alexnet}.
We extend the UFO$^2$ to the ResNet-50 FPN backbone to maintain consistency and compare it with our framework. Our P2BNet++-FR framework, an extension version of P2BNet-FR, outperforms Click and UFO$^2$ by 9.9 AP and 18.5 AP$_{50}$ on COCO-17. The P2MNet-FR framework also improves P2BNet++-FR’s performance to 25.9 AP on COCO-17, and with retraining on the latest detectors Sparse R-CNN and Swin-Transformer, the performance is increased to 29.6 AP and 30.3 AP, respectively. Furthermore, the results on COCO-14 were consistent. As shown in Fig.~\ref{fig:Vis}, the visualization demonstrates that the P2MNet-FR fully uses the point annotation’s precise location information and can distinguish dense objects in complex scenes. Compared with P2BNet++-FR, P2MNet-FR misses few objects and predicts tighter bounding boxes.

\noindent\textbf{Comparison with the Image-supervised methods.} 
We compare the proposed framework to the state-of-the-art Image-supervised methods on the COCO-14 dataset, and the results are presented in Table~\ref{tab:coco-tmp}. The experiments demonstrate that P2MNet-FR outperforms Image-supervised methods, and achieves significant improvements in detection performance with only a slight increase in annotation cost. These findings suggest that the Image-supervised methods task holds great promise for further development.

\noindent\textbf{Comparison with Box-Supervised methods.}
To validate the practicality of P2MNet-FR and demonstrate its performance upper bound under a supervised setting, we compare it with the box-supervised detectors in Table~\ref{tab:coco-tmp}. P2MNet-FR-R50 achieved an AP$_{50}$ of 49.0, which is much closer to the box-supervised detector FPN-R50 (58.1 AP$_{50}$) than the previous WSOD and PSOD methods. The results conducted on the latest detector show consistency. This indicates that PSOD can be used in industries~\cite{DBLP:ANTI-UAV,DBLP:2ND_ANTI-UAV} prioritizing object detection over the bounding box’s quality.

\subsubsection{Experiments on VOC}
The experiment verifies the effectiveness of our method on VOC 2007 and 2012 datasets. As shown in Table~\ref{tab:VOC_DET}, the FPN-R50 achieves 77.2 AP$_{50}$ and 75.3 AP$_{50}$ on VOC 2007, while the best WSOD method reaches 58.7 and 56.2 AP$_{50}$. In comparison, P2MNet-FR achieves 68.2 AP$_{50}$ and 66.7 AP$_{50}$ respectively, which significantly improves over the previous PSOD work. P2MNet-FR brings us much closer to fully supervised object detection. Moreover, the progressively improved performance of P2MNet-FR relative to P2BNet-FR and P2BNet++-FR illustrates the effectiveness of our method. The results of VOC 2012 also achieved consistently high performance.

\subsubsection{Experiments on Cityscapes}
We also inspect our method on the Cityscapes dataset to verify its generalizability, and the performance is presented in Table~\ref{tab:cityscape_OD}. Compared with the previous works, we follow the approach in~\cite{DBLP:MASK-RCNN} and report the performance with or without COCO pretraining in the detector. Specifically, we use either the pre-trained weights on COCO or ImageNet to initialize the network before training.  Since the Cityscapes dataset is relatively small, we pre-train P2BNet++ and P2MNet using the COCO weights by default. Our ImageNet or COCO pre-trained P2MNet-FR achieves 24.6 AP and 58.2 AP$_{50}$, and 27.0 AP and 58.2 AP$_{50}$, respectively. These results are close to FPN-R50, particularly on AP$_{50}$.

\begin{figure*}[tb!]
        \centering
        \includegraphics[width=1.\linewidth]{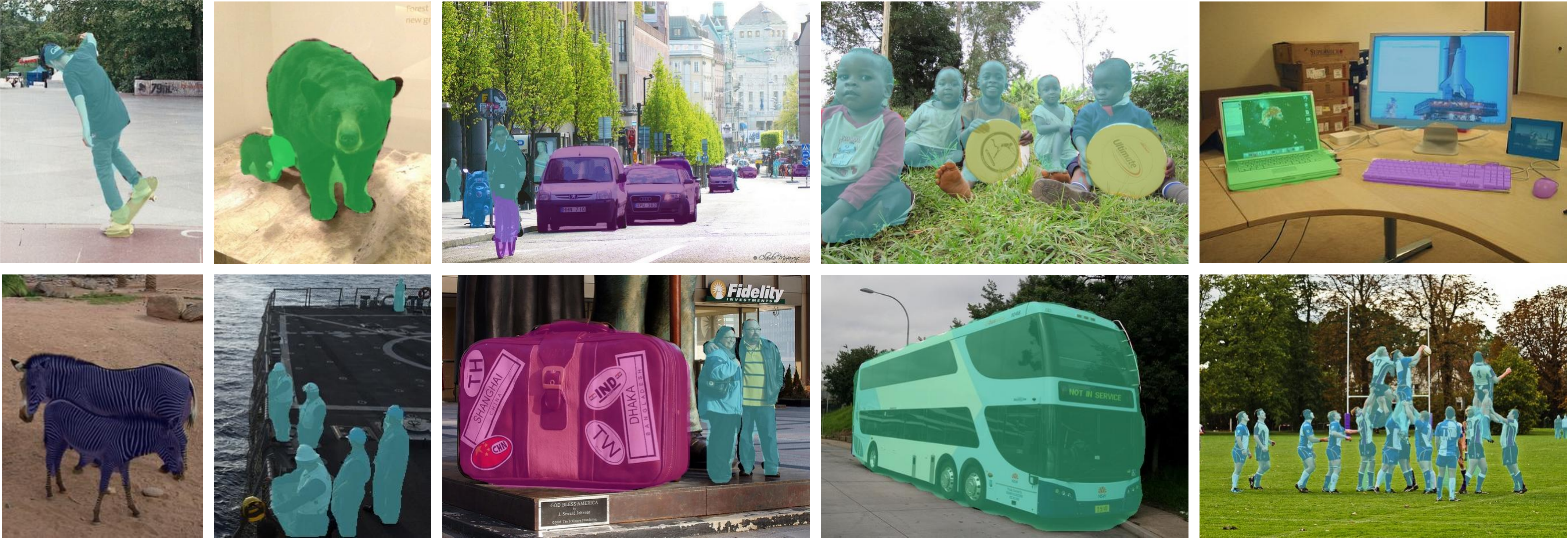}
    \caption{P2MNet-MR's instance segmentation results. Here, we show the results in different categories, different scales, and complex scenery. Only with single-point supervision, our segmentation results are surprising and show great prospects. (Best viewed in color.)}
    \vspace{-5pt}
    \label{fig:Vis_seg}
\end{figure*}

\begin{table}[tb!]
\small
\renewcommand\arraystretch{1.1}
\begin{center}
\resizebox{0.48\textwidth}{!}{
\begin{tabular}{l|ccccc}
\specialrule{0.13em}{0pt}{0pt}  
Method & Sup.& BB. &$\rm AP$ & $\rm AP_{50}$ & $\rm AP_{75}$ \\
\specialrule{0.08em}{0pt}{0pt}  
\color{gray} {Mask R-CNN~\cite{DBLP:MASK-RCNN}}&\color{gray} {$\mathcal{M}$} &\color{gray} {R-50}& \color{gray} {34.6} &\color{gray} {56.5}  &\color{gray} {36.6} \\
\color{gray} {CondInst~\cite{DBLP:condinst}} &\color{gray} {$\mathcal{M}$}&\color{gray} {R-50}&\color{gray} {35.3}& \color{gray} {56.4}& \color{gray} {37.4}\\
\color{gray} {SwinMR~\cite{DBLP:swin-transformer}} &\color{gray} {$\mathcal{M}$ }&\color{gray} {Sw-S}&\color{gray} {43.2}& \color{gray} {67.0} &\color{gray} {46.1} \\
\color{gray} {Mask2Former~\cite{DBLP:mask2former}} &\color{gray} {$\mathcal{M}$}&\color{gray} {Sw-S}& \color{gray} {46.1}&\color{gray} {69.4}&\color{gray} {52.8} \\

\color{gray} {BBTP~\cite{DBLP:BBTP}}& \color{gray} {$\mathcal{B}$}& \color{gray} {R-101}& \color{gray} {21.1}&\color{gray} {45.5}   &\color{gray} {17.2} \\
\color{gray} {BoxInst~\cite{DBLP:Boxinst}}& \color{gray} {$\mathcal{B}$}& \color{gray} {R-50}&\color{gray} {30.7} &\color{gray} {53.1}&\color{gray} {31.1}  \\

\color{gray} {IRNet~\cite{DBLP:IRNet}} &\color{gray} {$\mathcal{I}$}& \color{gray} {R-50}&\color{gray} {6.1} &\color{gray} {11.7}   &\color{gray} {5.5} \\
\color{gray} {BESTIE~\cite{DBLP:BESTIE}}  &\color{gray} {$\mathcal{I}$}  &  \color{gray} {HR-48} & \color{gray} {14.3}  & \color{gray} {28.0}  & \color{gray} {13.2} \\
 WISE-Net~\cite{DBLP:WISE-Net}&$\mathcal{P}$ &R-50&7.8  &18.2  & 8.8\\

BESTIE~\cite{DBLP:BESTIE}-MR$^{*}$&$\mathcal{P}$& HR-48 &17.7 & 34.0 &16.4 \\
Point2Mask~\cite{DBLP:point2mask}&$\mathcal{P}$ &R-101&12.8&26.3&11.2\\
Point2Mask~\cite{DBLP:point2mask}&$\mathcal{P}$ &Sw-L&14.6&29.5&13.0\\
AttnShift~\cite{DBLP:Attnshift}&$\mathcal{P}$ & Vit-S &19.1 &38.8&17.4 \\
AttnShift~\cite{DBLP:Attnshift}&$\mathcal{P}$ & Vit-B &21.2 &43.5&19.4 \\
SAM~\cite{DBLP:segment-anything}-MR&$\mathcal{P}'$& Vit-B+R-50&20.9&39.2&32.9\\
\specialrule{0.08em}{0pt}{0pt}  
\multicolumn{5}{l}{\textbf{\emph{Ours}}} \\
\specialrule{0.08em}{0pt}{0pt}  
\rowcolor{Gray}
   {P2BNet++-BoxInst}&  {$\mathcal{P}$}& {R-50}& {21.0} & {41.9}  &  {19.2} \\
 \rowcolor{Gray}

 {P2MNet-MR}&  {$\mathcal{P}$}& {R-50}& {23.1} &  {43.7}  &  {22.0} \\
 \rowcolor{Gray}
 {P2MNet-CondInst}&  {$\mathcal{P}$}&  {R-50}&  {23.3 }&  {43.7} &  {22.2} \\
 \rowcolor{Gray}
 {P2MNet-SwinMR$^{*}$}&  {$\mathcal{P}$}&  {R-50+Sw-S}& {26.2}&  {49.4} & {24.9}  \\
 \rowcolor{Gray}
 {P2MNet-M2Former$^{*}$}&  {$\mathcal{P}$}& {R-50+Sw-S}& {\textbf{28.3}} & {\textbf{50.2}}  & {\textbf{27.6}}  \\
 
\specialrule{0.13em}{0pt}{0pt}
\end{tabular}}
\end{center}

\caption{Instance segmentation performance of Mask-($\mathcal{M}$, N-points), image-($\mathcal{I}$, 0-point), box-($\mathcal{B}$, 4-points) and point-($\mathcal{P}$, 1-point) supervised methods on COCO-17 val. We mark SAM-MR with $\mathcal{P}'$ because it uses the foundation model SAM, which is trained on massive amounts of supervised data. So, it's not a pure point-supervised setting.
'Sup.' is the supervision type, and 'BB.' means backbone. $^{*}$ is multi-scale augment training for re-training segmentors, and other experiments are on single-scale training. SwinMR is Swin-Transformer - Mask R-CNN and M2Former is Mask2Former. HR-48 is the HRNet-48. SwinMR and M2Former use multi-scale.}
\vspace{-10pt}
\label{tab:wsiscoco2}
\end{table}

\subsection{Instance Segmentation Results}
\label{sec:PSIS}
We compare our P2MNet-MR framework against several existing instance segmentation methods, including those that rely on point supervision, image supervision, box supervision, and mask supervision. Additionally, we compare our framework to the latest advanced segmentation methods by altering the retrain network.
\subsubsection{Experiments on COCO}

\textbf{Comparison with the PSIS methods.} The previous PSIS methods have achieved the best performance of 21.2 AP on AttnShift with a strong backbone Vit-B, as shown in Table~\ref{tab:wsiscoco2}. BESTIE-HRNet-48 has achieved 17.7 AP. Point2Mask, a point-supervised panoptic segmentation method, achieves 12.8 and 14.6 AP with R-101 and Swin-L backbone. AttnShift achieves 21.2 AP with a Vit-B backbone, which is lower than ours (vs 23.1 AP for P2MNet-MR with R-50) but is based on a stronger backbone. We also use the newly proposed segment anything model (SAM) to generate proposal masks with our point annotation and choose the highest score mask as the pseudo mask to supervise Mask R-CNN (R50). It only achieves 20.9 AP, which is mainly because SAM is semantic-agnostic, which will bring ambiguity when only a single point is given.
Our P2BNet++ can obtain pseudo boxes, so we set up controlled experiments to train the BoxInst with pseudo boxes, and the performance is 21.0 AP. Our P2MNet-MR outperforms with 23.1 AP, demonstrating the effectiveness of our method. When retrained with the latest segmentation network and strong backbone Swin-S, we achieved 26.2 AP on Swin-Tranformer and 28.3 AP on Mask2Former, significantly improving the SOTA performance. Fig. 8 shows the impressive instance segmentation results.

\noindent\textbf{Comparison with the WSIS methods.}
The previous traditional WSIS methods, only supervised by image-level annotation, achieved unimpressive performance on the COCO dataset (Table~\ref{tab:wsiscoco2}). IRNet only achieves 6.1 AP, and BESTIE improves the performance to 14.3 AP.
Recently, BSIS, under box-level supervision, has achieved great improvement. The performance in BBTP is only 21.1 AP, while that in BoxInst rises to 30.7. With retraining on CondInst, the same structure as BoxInst, our method achieves 23.3 AP, close to the BSIS method. 


\begin{table}[tb!]
    \centering
    \resizebox{1.\linewidth}{!}{
    \begin{tabular}{l|ccccc}
    \specialrule{0.13em}{0pt}{0pt}
         Method&Sup.&Backbone&$\rm AP_{25}$&$\rm AP_{50}$&$\rm AP_{75}$\\
    \specialrule{0.08em}{0pt}{0pt} 
       \color{gray} {Mask R-CNN~\cite{DBLP:MASK-RCNN}}&\color{gray} {$\mathcal{M}$}&\color{gray} {R-50}&\color{gray} {78.0}&\color{gray} {68.8}&\color{gray} {43.3}\\
        \color{gray} {Mask R-CNN~\cite{DBLP:MASK-RCNN}}&\color{gray} {$\mathcal{M}$}&\color{gray} {R-101}&\color{gray} {79.6}&\color{gray} {70.2}&\color{gray} {45.3}\\
        
        \color{gray} {BoxInst~\cite{DBLP:Boxinst}} &\color{gray} {$\mathcal{B}$}&\color{gray} {R-101}&\color{gray} {-}&\color{gray} {61.4}&\color{gray} {37.0}\\
        \color{gray} {DiscoBox} &\color{gray} {$\mathcal{B}$}&\color{gray} {R-101}&\color{gray} {72.8}& \color{gray} {62.2} &\color{gray} {37.5} \\
        \color{gray} {BESTIE~\cite{DBLP:BESTIE}}& \color{gray} {$\mathcal{I}$} &\color{gray} {HR-48} &\color{gray} {53.5}&\color{gray} {41.7}&\color{gray} {24.2} \\
        \color{gray} {IRNet~\cite{DBLP:IRNet}} &\color{gray} {$\mathcal{I}$}&\color{gray} {R-50}&\color{gray} {-}&\color{gray} {46.7}&\color{gray} {23.5}\\
        \color{gray} {BESTIE~\cite{DBLP:BESTIE}}& \color{gray} {$\mathcal{I}$} &\color{gray} {HR-48 }&\color{gray} {61.2}&\color{gray} {51.0}&\color{gray} {26.6} \\
         WISE-Net~\cite{DBLP:WISE-Net}&$\mathcal{P}$&R-50&53.5&43.0&25.9 \\
         BESTIE~\cite{DBLP:BESTIE}&$\mathcal{P}$&HR-48&58.6&46.7&26.3 \\
         BESTIE~\cite{DBLP:BESTIE}-MR&$\mathcal{P}$&HR-48&66.4&56.1&30.2 \\
         Point2Mask~\cite{DBLP:point2mask}&$\mathcal{P}$ &R-101&-&48.4&22.8\\
            Point2Mask~\cite{DBLP:point2mask}&$\mathcal{P}$ &Sw-L&-&55.4&31.2\\
         
         Attnshift~\cite{DBLP:Attnshift} &$\mathcal{P}$&Vit-S&68.3 &54.4&25.4 \\
         Attnshift+imT~\cite{DBLP:Attnshift,DBLP:imTED} &$\mathcal{P}$&Vit-S&70.3 &57.1&30.4 \\
         \specialrule{0.08em}{0pt}{0pt}  
\multicolumn{5}{l}{\textbf{\emph{Ours}}} \\
\specialrule{0.08em}{0pt}{0pt}  
        \rowcolor{Gray}
         P2BNet++-BoxInst&$\mathcal{P}$&R-101 &67.5 &48.6&18.9\\
         \rowcolor{Gray}
         P2MNet-MR&$\mathcal{P}$&R-101 &\textbf{72.0}&\textbf{57.7}&26.1 \\
             \specialrule{0.13em}{0pt}{0pt}
    \end{tabular}}
\caption{Instance segmentation performance on SBD $test$ set. The segmentation network is trained on a multi-scale augment dataset using our method and previous networks for a fair comparison. imT is imTED~\cite{DBLP:imTED} described in~\cite{DBLP:Attnshift}. Here, `M' (Mask-level) can be viewed as polygons, we named it `N-points', `B' (Box-level) can be viewed as 4 points for supervision, we named it `4-points', `I' (Image-level) can be named as `0-point', `P' (Point-level) can be named as `1-point'.}
    \vspace{-10pt}
    \label{tab:voc_12_seg}
\end{table}

\begin{figure*}[tb!]
        \centering
        \includegraphics[width=1.\linewidth]{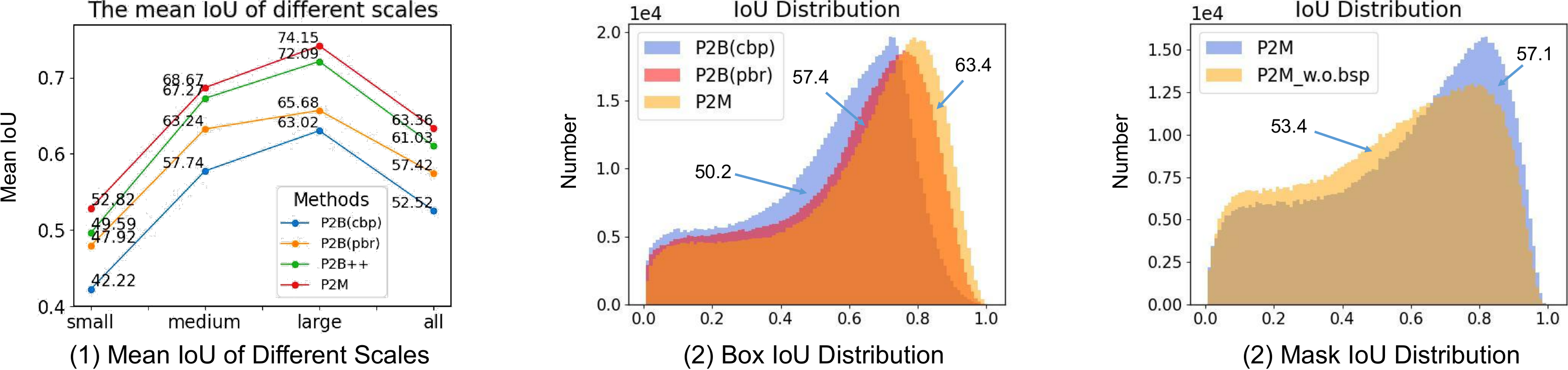}
    \vspace{-10pt}
    \caption{The statistics of the pseudo box/mask generation quality. (1) is the mIoU$_{box}$ (the mean IoU between predicted pseudo label and ground-truth label) of P2BNet(cbp), P2BNet(pbr), P2BNet++ and P2MNet on different scales. (2) and (3) are the pseudo box/mask IoU distributions for all objects. The number with an arrow is the mIoU$_{box/mask}$. They show that  BSP can help generate higher-IoU pseudo masks and P2MNet can improve the quality of pseudo boxes, which brings better detection performance.}
    \vspace{-0.5cm}
    \label{fig:stastic}
\end{figure*}

\noindent\textbf{Comparison with Mask-Supervised Methods.} As shown in Table~\ref{tab:wsiscoco2}, Mask R-CNN achieves 34.6 AP and 56.5 AP${50}$, while our P2MNet-MR achieves 23.1 AP and 43.7 AP${50}$. The latest segmentation model, Mask2Former with augmentation, achieves 46.1 AP and 69.4 AP${50}$, while our P2MNet-M2Former achieves 28.3 AP and 50.2 AP${50}$. Our method is approaching the performance of fully-supervised methods. However, there remains a performance gap. The method proposed in this paper can largely estimate the scale and shape information of the targets. However, in certain cases, the scale and shape are not accurately estimated, contributing to the performance gap. Additionally, while the approximate positions of the targets are predicted, some boundary details still require further refinement, which remains a challenge in the absence of supervision. We analyze the failure cases in Section~\ref{failure case}.
\begin{table}[tb!]
    \centering
    \resizebox{1.\linewidth}{!}{
    \begin{tabular}{l|cc|cc|cc}
    \specialrule{0.13em}{0pt}{0pt}
        \multirow{2}{*}{Method} &
        \multirow{2}{*}{\makebox[0.01\textwidth][c]{Sup.}}&  \multirow{2}{*}{Pretrain}& \multicolumn{2}{c|}{Detection} & \multicolumn{2}{c}{Segmentation}\\
         &&&$\rm AP$&$\rm AP_{50}$&$\rm AP$&$\rm AP_{50}$\\
    \specialrule{0.08em}{0pt}{0pt} 
       \color{gray} {FPN~\cite{FasterRCNN,DBLP:FPN}} &\color{gray} {$\mathcal{B}$}&\color{gray} {ImageNet }&\color{gray} {37.3} &\color{gray} {65.4}&\color{gray} {-}&\color{gray} {-} \\
       \color{gray} {FPN~\cite{FasterRCNN,DBLP:FPN}} &\color{gray} {$\mathcal{B}$}&\color{gray} {COCO} &\color{gray} {40.3} &\color{gray} {65.3} &\color{gray} {-}&\color{gray} {-}  \\
       \color{gray} {Mask R-CNN~\cite{DBLP:MASK-RCNN}} &\color{gray} {$\mathcal{B}$}&\color{gray} {ImageNet} &\color{gray} {39.4} &\color{gray} {66.0}&\color{gray} {34.2}&\color{gray} {61.0} \\
       \color{gray} {Mask R-CNN~\cite{DBLP:MASK-RCNN}} &\color{gray} {$\mathcal{B}$}&\color{gray} {COCO} & \color{gray} {40.9}&\color{gray} {66.2}&\color{gray} {36.4}&\color{gray} {61.8}\\
         UFO$^{2{\ddagger}}$~\cite{UFO2} &$\mathcal{P}$ &ImageNet&6.6&16.4&-&-\\
        UFO$^{2{\ddagger}}$~\cite{UFO2} &$\mathcal{P}$ &COCO&7.3&19.3&-&-\\
\specialrule{0.08em}{0pt}{0pt}  
\multicolumn{5}{l}{\textbf{\emph{Ours}}} \\
\specialrule{0.08em}{0pt}{0pt}  
        \rowcolor{Gray}
         P2BNet++-FR&$\mathcal{P}$&ImageNet &24.0&55.9&-&- \\
         \rowcolor{Gray}
         P2BNet++-FR&$\mathcal{P}$&COCO &26.9&59.5&-&- \\
         \rowcolor{Gray}
         P2MNet-FR&$\mathcal{P}$&ImageNet&25.0&58.1&-&-  \\
         \rowcolor{Gray}
         P2MNet-FR&$\mathcal{P}$&COCO &\textbf{27.4}&\textbf{58.7}&-&- \\
         \rowcolor{Gray}
         P2MNet-MR&$\mathcal{P}$&ImageNet&24.6&58.2&18.6&45.1  \\
         \rowcolor{Gray}
         P2MNet-MR&$\mathcal{P}$&COCO &27.0&58.2&\textbf{21.0}&\textbf{46.4} \\
    \specialrule{0.13em}{0pt}{0pt}        
    \end{tabular}}
    \caption{Object detection and instance segmentation results on Cityscapes $val$ set. The P2BNet++ and P2MNet are pre-trained on COCO by default. And the `pre-train' here means our FR or MR is pre-trained on ImageNet or COCO dataset. Here  `B' (Box-level) can be viewed as 4 points for supervision, we named it `4-points',  `P' (Point-level) can be named as `1-point'.}
    \vspace{-15pt}
    \label{tab:cityscape_OD}
\end{table}

\subsubsection{Experiments on SBD}
In Table.~\ref{tab:voc_12_seg}, AttnShift, the previous best PSIS method, achieves 57.1 AP$_{50}$. Our P2MNet-MR also exceeds it with 57.7 AP$_{50}$. The advantage of our method is the low missing rate, resulting in a higher AP$_{25}$. Compared to a fully supervised method, our AP$_{25}$ (72.0 vs. 79.6) and AP$_{50}$ (57.7 vs. 70.2) are comparable, indicating the prospects for practical applications.

\begin{figure*}
    \centering
    \vspace{-10pt}
    \includegraphics[width=0.85\linewidth]{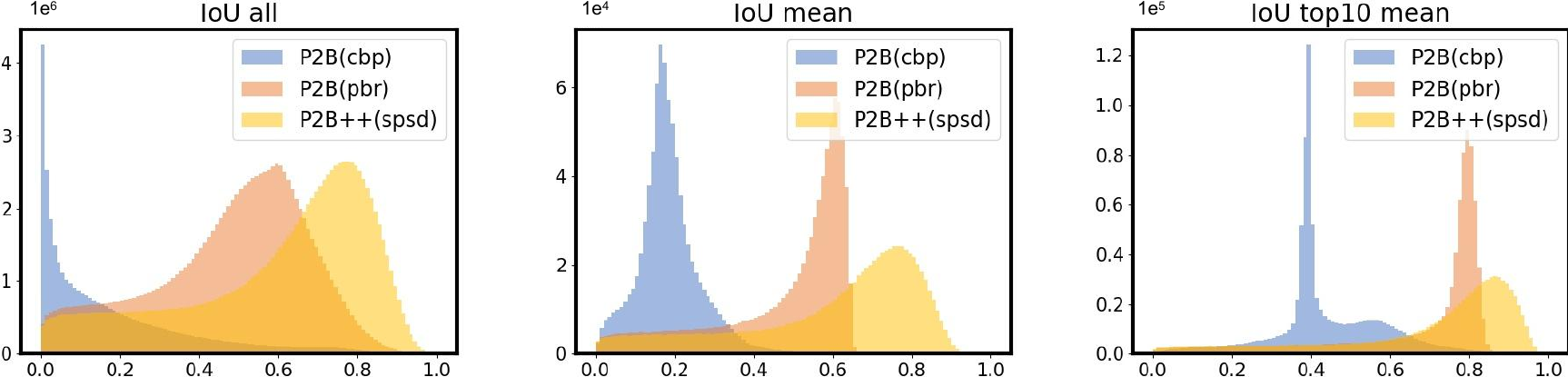}
    \caption{The statistics of proposals' IoU with GT. The `IoU all' means the distribution of IoU between proposals and GT, and the `IoU mean' means the distribution of average IoU between proposals and GT in each proposal bag.  `IoU all'  and `IoU mean' show the overall quality of proposal boxes. The `IoU top10 mean' means the distribution of average IoU between top-10 high-IoU proposals in each  bag and GT, showing the quality and proportion of high IoU proposals.}
    \label{fig:ioudist}
    \vspace{-15pt}
\end{figure*}

\begin{table*}[htb]
    \centering
    \resizebox{0.7\linewidth}{!}{
    \begin{tabular}{cc|cc|ccccccc}
    \specialrule{0.13em}{0pt}{1pt}
        \multicolumn{2}{c|}{CBP stage} & \multicolumn{2}{c|}{PBR stage} & \multicolumn{7}{c}{Performance}\\
        \specialrule{0.08em}{0pt}{1pt}
        $\mathcal{L}_{pos}$ &
        $\mathcal{L}_{mil1}$ & $\mathcal{L}_{mil2}$ & $\mathcal{L}_{neg}$  & mIoU$_{box}$& AP & AP$_{50}$& AP$_{75}$ &  AP$^s$ & AP$^m$ & AP$^l$\\
        \specialrule{0.08em}{0pt}{1pt}
         \checkmark & & \multicolumn{2}{c|}{} &25.0 &2.9 &10.3&0.7&1.1&3.1&5.1 \\
        & \checkmark  & \multicolumn{2}{c|}{}& 50.2 &13.7 & 37.8&6.8&5.8&14.3&21.4\\
        \specialrule{0.08em}{0pt}{1pt} 
        & \checkmark &\checkmark &  &52.0&12.7 &35.4&6.5&6.0&15.0&19.2\\
        \rowcolor{Gray}
        &  \checkmark& \checkmark&\checkmark&\textbf{57.4}&  \textbf{21.7}& \textbf{46.1}&\textbf{18.3}&\textbf{10.5}&\textbf{24.6}&\textbf{30.6}\\
        \specialrule{0.13em}{0pt}{0pt}
    \end{tabular}}
        \vspace{5pt}
        \caption{The effectiveness of the training loss in P2BNet: $\mathcal{L}_{mil1}$ in the CBP stage, $\mathcal{L}_{mil2}$ and $\mathcal{L}_{neg}$ in the PBR stage. mIoU$_{box}$ is the bounding box's mean IoU between pseudo boxes and GT.}
    \vspace{-10pt}
    \label{tab:main loss}
\end{table*}

\begin{table}[tb!]
\centering
\resizebox{\linewidth}{!}{
\begin{tabular}{lcccc}
\specialrule{0.13em}{0pt}{0pt}
Method & mIoU & mIoU$_s$& mIoU$_m$ & mIoU$_l$ \\
\specialrule{0.08em}{0pt}{0pt}
CBP & 50.2 & 39.8 & 54.9 & 59.4 \\
CBP(w. PBR jointly)  & 52.5 & 42.2 & 56.5 & 62.1 \\
\specialrule{0.13em}{0pt}{0pt}
\end{tabular}}
\caption{The quality of estimated pseudo box (CBP vs CBP with PBR jointly optimization). This shows that CBP and PBR stages are not decoupled. PBR helps CBP learn better, avoiding sub-optimality.}
\vspace{-15pt}
\label{CBP-joint}
\end{table}

\subsubsection{Experiments on Cityscapes}
PSIS's performance is shown in Table.~\ref{tab:cityscape_OD}. The mask-supervised Mask R-CNN is 34.2 AP and 36.4 AP with the ImageNet or COCO pretrain. Our PSIS method P2MNet-MR achieves 18.6 AP and 45.1 AP$_{50}$ on the Cityscapes validation set. With the pre-trained weights on the COCO dataset, the performance increases to 21.0 AP and 46.4 AP$_{50}$.

\subsection{Ablation Study of P2BNet}
This section presents all the ablation studies conducted on the COCO-17 dataset. The top-$k$ setting is $k=7$ except for  Table~\ref{tab:top k}. 

\label{sec:abla_p2mnet}

\noindent\textbf{Training loss in P2BNet.} Table~\ref{tab:main loss} shows the ablation study of the training loss in P2BNet. \textbf{i) CBP loss.} By using only $\mathcal{L}_{mil1}$ in the CBP stage, we can obtain 13.7 AP. For comparison, we conduct $\mathcal{L}_{pos}$, which views all the proposal boxes in the bag as positive samples. However, we found it hard to optimize, and the performance was unsatisfactory. Coarse proposal bags can cover most objects in the high IoU, resulting in a low missing rate. The performance still has the potential to be gained by setting denser anchors.
\noindent\textbf{ii) PBR loss.} The PBR sampling is based on the coarse pseudo-box output from CBP and performs fine scaling, as well as jittering of scale, aspect ratio, and center position. This forms a coarse-to-fine cascading process that increases the upper limit for proposal selection, as shown in Fig.~\ref{fig:ioudist}. The performance is only 12.7 AP with $\mathcal{L}{mil2}$ in PBR because of error accumulation in a cascade fashion and a lack of negative samples for focal loss. 
To address this issue, considering that there are no explicit negative samples to suppress the $Sigmoid$ activation function background, we introduce a negative sampling and a negative loss $\mathcal{L}{neg}$. Due to the approximate nature of CBP's predictions, regions far from the pseudo-box can be sampled as negative examples.
As a result, the performance increases by 9.0 AP.
We also evaluate the mIoU$_{box}$ to discuss the predicted pseudo box's quality. 
In the PBR stage with $\mathcal{L}_{mil2}$ and $\mathcal{L}_{neg}$, the mIoU increases from 50.2 to 57.4, suggesting a better quality of the pseudo box. 
The optimization of PBR and CBP is jointly trained, with both sharing the backbone and two fully connected (FC) layers. PBR's high-quality positive samples and background suppression through negative samples aid in CBP's learning and help prevent sub-optimal solutions. Table~\ref{CBP-joint} shows that when CBP is trained independently, the quality of the estimated pseudo boxes is 50.2 mIoU. However, when trained jointly with PBR, the predicted pseudo boxes from CBP reach 52.5 mIoU. This demonstrates that PBR optimization helps CBP learn more effectively.

\begin{table}[tb!]
    \centering
    \resizebox{1.\linewidth}{!}{
        \begin{tabular}{c|ccccccc}
        \specialrule{0.13em}{0pt}{1pt}
        top-$k$ &mIoU$_{box}$ &AP &AP$_{50}$ & AP$_{75}$ &  AP$^s$ & AP$^m$ & AP$^l$\\
        \specialrule{0.08em}{0pt}{1pt}
             1& 49.2 &12.2 &35.9&4.9&9.3&14.9&15.1 \\
             3& 54.7 &21.3 & 46.6&16.7&11.9&24.0&28.7\\
                     \rowcolor{Gray}
             4& \textbf{57.5} & \textbf{22.1} & \textbf{47.3}&18.1 &\textbf{11.5}&\textbf{24.8}&30.4\\
             7& 57.4 & 21.7 & 46.1&\textbf{18.3}&10.5&24.6&\textbf{30.6}\\ 
             10& 57.1 & 21.5 & 46.0&17.5&10.2&24.4&30.7\\
        \specialrule{0.13em}{0pt}{0pt}
        \end{tabular}}
        \caption{The top-$k$ policy for box merging. 
        }
        \vspace{-15pt}
        \label{tab:top k}
\end{table}

\begin{table}[tb!]
    \centering
    \resizebox{1.\linewidth}{!}{
    \begin{tabular}{l|ccccc}
    \specialrule{0.13em}{0pt}{1pt}
    Methods & mIoU$_{box}$& AP &AP$_{50}$& AP$_{75}$ \\
    \specialrule{0.08em}{0pt}{1pt}
    One-stage P2BNet (CBP)&50.2  &13.7 & 37.8&6.8\\
         Two-stage P2BNet&57.4 &21.7 &46.1&18.3  \\
         Three-stage P2BNet&57.0&21.9&46.1&18.2 \\
         \rowcolor{Gray}
         Three-stage P2BNet $w.$ SPSD
         &\textbf{61.0}& \textbf{23.1} &\textbf{47.4} &\textbf{19.9} \\
    \specialrule{0.13em}{0pt}{1pt}
    \end{tabular}}
        \caption{Effectiveness of SPSD in P2BNet++.}
    \vspace{-15pt}
    \label{tab:P2BNet+}
\end{table}

\begin{table*}[htb!]
    \centering
    \resizebox{1.\linewidth}{!}{
    \begin{tabular}{l|ccccccc|ccccccc}
        \specialrule{0.13em}{0pt}{1pt}
        \multirow{2}{*}{Methods} & \multicolumn{7}{c|}{Object Detection (+FR)} & \multicolumn{7}{c}{Instance Segmentation (+MR)} \\
        & mIoU$_{box}$ & AP &AP$_{50}$&AP$_{75}$& AP$^s$&AP$^m$& AP$^l$& mIoU$_{mask}$ & AP &AP$_{50}$& AP$_{75}$&AP$^s$&AP$^m$& AP$^l$\\
         \specialrule{0.08em}{0pt}{1pt}
        P2BNet++& 61.0&23.1&47.4&19.9&11.0& 25.4 &32.9&-&-&-&-&-&-&-   \\
        $w.$ OAPP &60.8&23.3&48.1&20.3&11.2&25.5&33.6 &53.4&21.1&42.3&19.2&7.1&23.0&35.3\\
        \rowcolor{Gray}
        $w.$ OAPP $\&$ BSP  &\textbf{63.4}&\textbf{25.9}&\textbf{49.0}&\textbf{24.6}&\textbf{12.9}&\textbf{29.0}&\textbf{36.0}&\textbf{57.1}&\textbf{23.1}&\textbf{43.7}&\textbf{22.0}&\textbf{9.4}&\textbf{25.4}&\textbf{35.9}\\
         \specialrule{0.13em}{0pt}{1pt}
    \end{tabular}}
    \caption{The ablation of OAPP and BSP modules for Object detection and Instance segmentation.} 
    \vspace{-10pt}
     \label{tab:ModulesofP2MNet}
\end{table*}

\noindent\textbf{Box merging policy.}  We use the Top-$k$ score average weight as the merging policy. The hyper-parameter $k$ is slightly sensitive and easily generalized to other datasets, as presented in Table~\ref{tab:top k}. Only the top-$1$ or top-$few$ proposal box plays a leading role in box merging. The best performance is 22.1 AP and 47.3 AP$_{50}$ when $k=4$. 
The mIoU$_{box}$ for the preicted box is 57.5. During inference, if we only use the classification score $\mathbf{S}^{cls}$ instead of the bag score $\mathbf{S}$
for merging, the performance drops to 17.4 AP (vs 21.7 AP).

\noindent\textbf{SPSD sampling in P2BNet++.}
\label{sec:p2b++_analysis}
P2BNet's proposal construction relies on discrete, manually set anchor sampling. Denser anchor settings (basic PBR sampling) in the PBR stage improve performance (an 8.0 AP gain, reaching 21.7 AP), but they still do not yield an optimal solution, with performance saturating despite further refinements. We investigate the effect of SPSD sampling and present the results in Table~\ref{tab:P2BNet+}.
P2BNet++ leverages spatial information and uses pseudo-boxes to supervise proposal box regression, thereby improving the quality and proportion of high IoU proposal boxes (as shown in Fig.~\ref{fig:ioudist}), which, in turn, helps increase the upper limit of MIL selection. We added a PBR stage with the SPSD sampling after P2BNet. For a fair comparison, we set a three-stage experiment (one CBP stage and two PBR stages) with both the original P2BNet and P2BNet with SPSD sampling individually. We observe that P2BNet with SPSD sampling shows a significant increase of 1.2 points, indicating the effectiveness of the SPSD sampling. We entitle the three-stage P2BNet with SPSD sampling in the last PBR stage as P2BNet++.

\subsection{Ablation Study of P2MNet}

\noindent\textbf{Modules in P2MNet.}
The ablation study of the proposed modules (with their corresponding loss function) in P2MNet is shown in Table~\ref{tab:ModulesofP2MNet}. The first line shows P2BNet++'s performance, which serves as our detection baseline for P2MNet. With OAPP, P2MNet predicts objects' mask representation and uses the mask's minimum external rectangle to supervise Faster R-CNN. This results in some improvement, but it is limited due to the imprecise supervision by the bounding box. Through in-depth analysis, it is found that the image color similarity constraint is related only to the image itself, making the design of the pixel similarity loss in mask prediction reasonable. However, the boundary contour details estimated by P2BNet (or P2BNet++) are not always reliable. BSP is designed to relax the projection constraint at the target boundary, allowing the boundary to self-correct based on the network’s learned statistical probability within the approximate pseudo-box range. As shown in the third row of Table 10, BSP leads to improved mIoU scores for both bounding boxes and masks. The detection and segmentation performances increase to 25.9 AP and 23.1 AP, with improvements of 2.2 and 1.1 points, respectively. Fig.~\ref{fig:Vis_analysis} illustrates that with the BSP module, the object boundary becomes smoother, and prediction errors such as truncation and over-inclusion are alleviated. These results demonstrate that P2MNet generates more precise pseudo bounding boxes and high-quality pseudo masks, as shown in Fig.~\ref{fig:stastic}, ultimately leading to better detection and segmentation performances.

\noindent\textbf{Region rate of BSP.}
We conduct an investigation in P2MNet on the impact of the BSP module's region rate in Table~\ref{tab:region rate}. The results indicate that the optimal performance is achieved when the region rate is set to 0.1. However, the performance drops if the region rate is set too high or too low. The mIoU$_{box}$ and mIoU$_{mask}$ reach 63.4 and 57.1, and the detection and segmentation performances achieve 25.9 AP and 23.1 AP. 

\begin{table}[t!]
    \centering
    \resizebox{1.\linewidth}{!}{
    \begin{tabular}{l|ccc|ccc}
    \specialrule{0.13em}{0pt}{1pt}
    \multirow{2}{*}{Rate} &  \multicolumn{3}{c|}{Object Detection (+FR)} & \multicolumn{3}{c}{Instance Segmentation (+MR)} \\
    &mIoU$_{box}$& AP &AP$_{50}$& mIoU$_{mask}$&AP &AP$_{50}$ \\
    \specialrule{0.08em}{0pt}{1pt}
    -&60.8&23.3&48.1&53.4&21.1&42.3\\
    0.05&62.5&25.4&48.7&55.6&22.7&43.3 \\
    \rowcolor{Gray}
    0.1&\textbf{63.4}&\textbf{25.9}&\textbf{49.0}&57.1&\textbf{23.1}&\textbf{43.7} \\
    0.2&62.6&24.3&48.5&\textbf{57.4}&22.0&43.5\\

    \specialrule{0.13em}{0pt}{1pt}
    \end{tabular}}
    \caption{Effect of region rate of BSP in P2MNet.} 
    \vspace{-15pt}
    \label{tab:region rate}
\end{table}

\begin{table}[t!]
    \centering
    \resizebox{1.\linewidth}{!}{
    \begin{tabular}{l|ccc|ccc}
    \specialrule{0.13em}{0pt}{1pt}
    \multirow{2}{*}{Setting} &  \multicolumn{3}{c|}{Object Detection (+MR)} & \multicolumn{3}{c}{Instance Segmentation (+MR)} \\
    &mIoU$_{box}$& AP &AP$_{50}$& mIoU$_{mask}$&AP &AP$_{50}$ \\
    \specialrule{0.08em}{0pt}{1pt}
    -&60.8&23.3&47.9&53.4&21.1&42.3\\
    inner&62.7&24.4&\textbf{48.4}&56.8&22.4&\textbf{44.0}\\
    outer&60.6&23.1&47.4&54.4&22.0&42.8\\
    \rowcolor{Gray}
    both&\textbf{63.4}&\textbf{25.6}&\textbf{48.4}&\textbf{57.1}&\textbf{23.1}&43.7 \\
    \specialrule{0.13em}{0pt}{1pt}
    \end{tabular}}
    \caption{Region type of BSP in P2MNet. The results are all based on Mask R-CNN.}
    \vspace{-15pt}
    \label{tab:Region type}
\end{table}

\begin{table}[t]
    \centering
    \resizebox{1.\linewidth}{!}{
    \begin{tabular}{l|cc|cc|cc}
    \specialrule{0.13em}{0pt}{0pt}
    \multirow{2}{*}{Detectors} &  \multicolumn{2}{c|}{GT box} & \multicolumn{2}{c|}{P2BNet box}&
  \multicolumn{2}{c}{\cellcolor{Gray}P2MNet box}\\
    &AP & AP$_{50}$&AP & AP$_{50}$&  \cellcolor{Gray}AP &   \cellcolor{Gray}AP$_{50}$\\
    \specialrule{0.08em}{0pt}{0pt}
    RetinaNet~\cite{DBLP:retinanet_focalloss} &36.5&55.4 &21.0 &44.9&   \cellcolor{Gray}25.4&  \cellcolor{Gray}47.6\\
    Reppoint~\cite{DBLP:reppoint} &37.0 & 56.7 &20.8 &45.1&  \cellcolor{Gray}24.9&  \cellcolor{Gray}47.0 \\
    FR-FPN~\cite{FasterRCNN,DBLP:FPN} &37.4&58.1&22.1 &47.3 &  \cellcolor{Gray}25.9&  \cellcolor{Gray}49.0\\
    Sparse R-CNN~\cite{DBLP:sparsercnn} &\textbf{45.0} &\textbf{64.1} &\textbf{24.6}&\textbf{49.6}&  \cellcolor{Gray}\textbf{29.6}&  \cellcolor{Gray}\textbf{52.5}\\
    \specialrule{0.13em}{0pt}{0pt}
    \end{tabular}}
    \caption{Performance of different detectors on GT box annotations and pseudo boxes generated by P2BNet and P2MNet. Sparse R-CNN uses augment and 3x schedule.}
    \vspace{-15pt}
    \label{tab:detectors}
\end{table}

\noindent\textbf{Region type of BSP.}
We set the default mode of the BSP module to self-predict both the inner-side and outer-side regions of the bounding box. 
In our study, we investigated the impact of using either the inner-side or outer-side regions, and the results are shown in Table~\ref{tab:Region type}. We find that using BSP on either side alone is less effective than on both sides.
Specifically, the outer-side BSP helps to solve the truncated prediction problem but causes more noise. In contrast, the inner-side BSP helps reduce the overinclusive prediction but results in more serious truncation problems.

\noindent\textbf{Different detectors.}
To conduct the integrity experiments, we train different detectors~\cite{FasterRCNN,DBLP:FPN,DBLP:retinanet_focalloss,DBLP:reppoint,DBLP:sparsercnn} on R-50, as shown in Table~\ref{tab:detectors}. Our framework demonstrates competitive performance on all these detectors. Moreover, the detectors trained with pseudo boxes that are generated by P2MNet consistently outperform the others, showing the generalizability of our approach across different detectors. 
We list the box-supervised performances to highlight the upper bound of our framework.

\begin{table}[tb!]
    \centering
       \resizebox{1.\linewidth}{!}{
\begin{tabular}{c|c|c|c|c}
    \specialrule{0.13em}{0pt}{0pt}
Type & Image tag & Point & Box & Mask \\
    \specialrule{0.08em}{0pt}{0pt}
\multirow{2}{*}{Cost} & 1.5 s/img & 1.87 s/img & 34.5 s/img & 239.7 s/img \\
& - & 0.9 s/obj & 7 s/obj & 79.2 s/obj \\
    \specialrule{0.13em}{0pt}{0pt}
\end{tabular}}
    \caption{The cost of different annotation types. `s' , `img' and `obj' are `second',  `image' and `obj', respectively. }
    \vspace{-10pt}
    \label{tab:ann cost}
\end{table}

\subsection{Failure cases}
\label{failure case}

\begin{figure}
    \centering
    \includegraphics[width=1.0\linewidth]{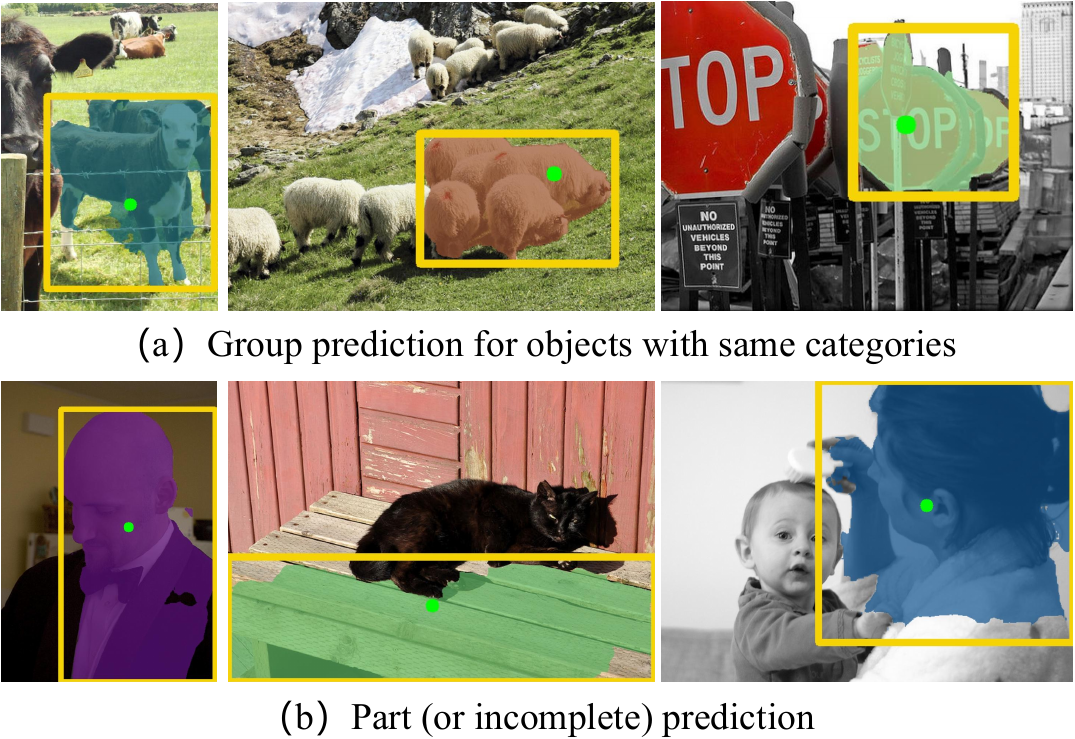}
    \vspace{-10pt}
    \caption{Analysis of the failure cases: In the first instance, it is evident that the issue of adjacent targets belonging to the same category poses a significant challenge for point-supervised methods. As for the second aspect, it demonstrates that relying solely on point annotation leads to the estimation being concentrated on parts, consequently resulting in incomplete predictions.}
    \vspace{-10pt}
    \label{fig:bad case}
\end{figure}

We present the failure cases in Figure \ref{fig:bad case}, which highlight issues related to group prediction and partial (or incomplete) prediction. The former occurs when objects of the same category are adjacent, causing confusion for the network. MIL-based methods rely solely on classification ability, and since objects of the same category all have high scores, the network may select a region containing multiple objects as the output. The latter issue arises when part of the object is discriminated against and receives a higher score, while the goal is to detect the entire object. Additionally, pixel similarity may lead the network to determine object boundaries based on color texture rather than semantics, such as in the case of a white shirt on a person’s body. This results in incomplete predictions. Through these failure cases, we aim to provide valuable insights to the community and will continue to explore these challenges further.

\subsection{Discussion}
\subsubsection{Advantages and limitations of Point-Level Supervision}
In this subsection, we discuss the advantages and limitations of point-level supervision, highlighting its value and potential directions for future improvement.

\noindent\textbf{(1) Advantages.}

\noindent\noindent\textbf{Better performance than traditional weakly-supervised methods:} As shown in Tables~\ref{tab:coco-tmp} and~\ref{tab:voc_12_seg}, the proposed method outperforms traditional weakly-supervised methods, even surpassing the state-of-the-art (SOTA) in some cases. This demonstrates its promising potential for practical applications.

\noindent\noindent\textbf{Lower annotation cost than traditional fully-supervised methods:} The average time for point annotation is approximately 1.87 seconds per image, as shown in Table~\ref{tab:ann cost}, which is comparable to image-tag annotation (1.5 s/img). The statistics~\cite{DBLP:C-WSL,Click} are based on VOC~\cite{DBLP:VOC}, and a similar approach can be applied to COCO~\cite{coco}. In terms of instance-level annotation time~\cite{DBLP:pointly-sup,DBLP:Extreme_Clicking,coco} for COCO, point annotation (0.9 s/obj) is significantly faster than bounding box annotation (7 s/obj) and polygon-based mask annotation (79.2 s/obj). Point annotation serves as a data supplement for object detection in specific domains. While models trained on large-scale datasets generally perform well in broad scenarios, they may underperform in specific domains. In such cases, low-cost and efficient data are essential for quick fine-tuning. This highlights the significance of point-level annotation.

\noindent\textbf{(2) Limitations.} 

\noindent In dense scenarios, point-level supervision saves annotation time compared to traditional full supervision. However, it can still be labor-intensive, as many objects need to be clicked. Therefore, combining point-level supervision with few-shot learning methods is a promising research direction. For example, annotating a few people on a beach could allow the model to automatically annotate all people in the scene.

\subsubsection{Practical Implications.}

\noindent\textbf{Data Supplement.} Our method can be used to construct datasets by converting point annotations into pseudo box labels or pseudo mask labels for detection and segmentation tasks. The value of point supervision lies in its ability to rapidly and cost-effectively generate annotated data, helping models in specific domains. This enables detection or segmentation models to be transferred to new target areas at a low cost. In real-world scenarios, particularly in specialized domains, lightweight point-level annotation is more cost-effective and advantageous for handling large data volumes, serving as a significant data supplement.

\noindent
\noindent\textbf{Model Deployment.} Furthermore, during the inference stage, only the trained detector or segmentor is employed. P2BNet (or P2MNet) is used solely to generate pseudo labels during the training phase. As a result, there is no additional computational burden during inference, and the trained detector can be directly deployed for practical applications.

\section{Conclusion}
In this paper, propose a novel P2Object framework, containing P2BNet and P2MNet, that predicts the object bounding box or mask with a single point. P2BNet generates inter-objects balanced and high-quality instance-level proposal bags for MIL training. A fixed anchor-like sampling is employed around the point and instance-level MIL is utilized to predict coarse pseudo boxes. The cascaded architecture and spatial self-distillation sampling strategy refine the pseudo boxes. P2BNet++ conducts an approximately continuous proposal sampling strategy by
better utilizing spatial clues. Through further study, we incorporate a BSP module, combining low-level pixel information, to alleviate the truncated and overinclusive prediction issues caused by discrete box sampling, making the optimization continuous, yielding P2MNet. This transfers discrete spatial sampling into the continuous pixel-level object perception. Additionally, pixel prediction helps to perceive objects more accurately and extends the framework into the instance segmentation task.
Remarkably, the conceptually simple framework delivers state-of-the-art object detection and instance segmentation performance with a single-point annotation across multiple datasets.

\noindent\textbf{Future and prospects.} P2BNet now has been widely generalised to aerial orientated object detection~\cite{DBLP:pointobb,DBLP:point2rbox,DBLP:p2rbox,DBLP:p-to-rbox,DBLP:pointobb-v2}, noisy-supervised object detection~\cite{DBLP:SSD-Det} tasks. Therefore, exploiting more applications of P2BNet and P2MNet in point-supervised tasks can be further studied. In orientated object detection, the orientated anchors can be sampled as a bag to estimate angles through MIL training. In the noisy-supervised detection task, the PBR stage can be seen as a noisy box refinement policy.
 In addition, the combination of our method and the visual foundation model, \eg~SAM, captures our interest. The predicted boxes or masks can be utilized as the interactive prompt of SAM to generate more precise masks.

 \section{Data Availability Statements}
We use the four publicly available datasets in our paper: MS COCO~\cite{coco},  Pascal VOC~\cite{DBLP:VOC}, SBD~\cite{DBLP:VOC} and Cityscapes~\cite{cityscape}. MS COCO is available at \url{https://cocodataset.org/}. Pascal VOC and SBD are available at \url{http://host.robots.ox.ac.uk/pascal/VOC/}. Cityscapes is available at \url{227。9https://www.cityscapes-dataset.com/}.

\bibliographystyle{spbasic}
\bibliography{main}

\end{document}